\documentclass[10pt]{article} % For LaTeX2e

\usepackage[accepted]{rlj} % Should be uncommented for the camera-ready
\usepackage{amsmath,amsfonts,bm}

\def\eqref#1{equation~\ref{#1}}
\def\1{\bm{1}}

\DeclareMathAlphabet{\mathsfit}{\encodingdefault}{\sfdefault}{m}{sl}
\SetMathAlphabet{\mathsfit}{bold}{\encodingdefault}{\sfdefault}{bx}{n}

\usepackage{amssymb}            % Defines common symbols like \mathbb R
\usepackage{mathtools}          % Extends amsmath, providing common math tools
\usepackage{mathrsfs}           % Enables \mathscr, which can work in cases that \mathcal does not
\usepackage{graphicx}           % For including images
\usepackage{subcaption}         % Allows for the use of subfigures and subcaptions
\usepackage[space]{grffile}     % For spaces in image names
\usepackage[frozencache, cachedir=.]{minted2}
\usepackage{booktabs}           % For high-quality tables
\usepackage{algorithm}          % For algorithm environments
\usepackage{algpseudocode}      % For algorithmicx pseudocode
\usepackage{appendix}           % For appendix formatting
\usepackage[capitalize]{cleveref} % For smart cross-referencing
\usepackage{inconsolata}

\hypersetup{
  colorlinks   = true,              %Colours links in   stead of ugly boxes
  urlcolor     = blue,              %Colour for external hyperlinks
  linkcolor    = blue,              %Colour of internal links
  citecolor    = blue                %Colour of citations
}
\title{When Do We Need LLMs? \\A Diagnostic for Language-Driven Bandits}
\setrunningtitle{A Diagnostic for Language-Driven Bandits}

\author{Uljad Berdica\textsuperscript{1,$*$}, Fernando Acero\textsuperscript{2}, Anton Ipsen\textsuperscript{2}, Parisa Zehtabi\textsuperscript{2}, \\ Michael Cashmore\textsuperscript{2}, Manuela Veloso\textsuperscript{2}}

\emails{uljadb@robots.ox.ac.uk , \{firstname.lastname\}@jpmorgan.com}
\affiliations{
$^{1}$\textbf{University of Oxford}, 
$^{2}$\textbf{J.P. Morgan AI Research}
\par % If including additional comments like below, use \par to add some whitespace. 
$^*$ Work done during an internship at J.P. Morgan.
}
\contribution{
    We introduce LLM Process UCB (LLMP-UCB), a novel bandit algorithm that combines the semantic reasoning capabilities of Large Language Models with the principled uncertainty estimation of UCB-style numerical bandits.
    }
    {
    Existing LLM-based bandit methods either lack reliable uncertainty estimates or require extensive prompt engineering~\citep{monea2025llmsincontextbanditreinforcement,sun2025largelanguagemodelenhancedmultiarmed} or fine-tuning~\citep{schmied2025llms} to induce exploration. In contrast, LLMP-UCB adapts the LLM Process framework~\citep{requeima2024llmprocessesnumericalpredictive} to sequential decision-making, obtaining calibrated reward distributions without architectural modifications to the underlying model.
    }

\contribution{
    We empirically demonstrate that lightweight numerical contextual bandits operating directly on text embeddings (dense or Matryoshka) frequently match or exceed the performance of LLM-based agents at a fraction of the computational cost.
    }
    {
    Across synthetic and real-world classification tasks, we find that deploying LLMs is only necessary when reward functions are highly non-linear or defined by complex semantic criteria. For many practical settings, a simple LinUCB on dense or Matryoshka embeddings is sufficient. These insights are helpful in restricting the deployment of LLMs to settings where their comparatively higher cost is justified by higher performance.
    }

\contribution{
    We show that the dimensionality of the text embedding space directly controls the exploration-exploitation tradeoff, and propose a geometric diagnostic to guide practitioners in choosing between LLM-driven and embedding-based approaches.
    }
    {
    Lower-dimensional embeddings accelerate learning but plateau at lower accuracy; higher dimensions achieve better asymptotic performance at greater sample cost. This provides a tunable, interpretable lever for cost–performance optimization that avoids reliance on prompt engineering.
    }

 % Your keywords

\summary{Decision-making systems increasingly encounter contexts that blend numerical data with natural language, such as recommendation systems or portfolio management based on financial reports. Large Language Models (LLMs) offer a natural interface for such problems, but deploying them at every decision step is expensive and yields poorly calibrated uncertainty estimates. This work addresses the main question of when LLMs are actually necessary for language-driven bandit problems or whether a numerical is more suitable. We introduce LLMP-UCB, which extracts distributional uncertainty from LLMs via repeated inference, and systematically compare it against lightweight numerical bandits operating on text embeddings. Following our evaluation of different instances of bandit architectures, we make the following practical findings: embedding-based methods often suffice, embedding dimensionality provides a direct lever on exploration–-exploitation tradeoffs, and we offer a geometric diagnostic to help practitioners choose the right tool for their problem.}

\begin{document}

% \makeCover  % Create the cover page
\maketitle  % Make the title section

\begin{abstract}
We study Contextual Multi-Armed Bandits (CMABs) for non-episodic decision-making problems where the context includes both textual and numerical information (e.g., recommendation systems, dynamic portfolio adjustments, offer selection; all frequent problems in finance). While Large Language Models (LLMs) are increasingly applied to these settings, utilizing LLMs for reasoning at every decision step is computationally expensive, and uncertainty estimates are difficult to obtain. To address this, we introduce LLMP-UCB, a bandit algorithm that derives uncertainty estimates from LLMs via repeated inference. However, our experiments demonstrate that lightweight numerical bandits operating on text embeddings (dense or Matryoshka) match or exceed the accuracy of LLM-based solutions at a fraction of their cost. We further show that embedding dimensionality is a practical lever on the exploration-exploitation balance, enabling cost–performance tradeoffs without prompt complexity. Finally, to guide practitioners, we propose a geometric diagnostic based on the arms' embeddings to decide when to use LLM-driven reasoning versus a lightweight numerical bandit. Our results provide a principled deployment framework for cost-effective, uncertainty-aware decision systems with broad applicability across AI use cases.
\end{abstract}

%%%%%%%%%%%%%%%%%%%%%%%%%%%%%%%%%%%%%%%%%%%%%%%%%%%%%%%%%%%%%%%%
%% Section: Submission of papers to RLJ/RLC
%%%%%%%%%%%%%%%%%%%%%%%%%%%%%%%%%%%%%%%%%%%%%%%%%%%%%%%%%%%%%%%%
\section{Introduction}

Contextual Multi-Armed Bandits (CMABs) provide a framework for sequential decision-making in which an agent selects actions based on observed context and receives stochastic rewards~\citep{lattimore2020bandit, Sutton1998}. Unlike full reinforcement learning, CMABs do not require modeling state transitions, making them well-suited to problems where interactions are independently sampled rather than temporally dependent. This simplicity has enabled broad adoption across domains including recommendation systems~\citep{li2010contextual}, clinical trials~\citep{villar2015multi}, and hyperparameter optimization~\citep{li2018hyperband}. However, standard CMAB algorithms operate on numerical feature vectors and cannot directly incorporate natural language. This limitation is significant for applications where textual information (such as product descriptions, user reviews, or document content) is central to the decision problem.

Large Language Models (LLMs) have emerged as a promising solution for directly applying language to decision-making problems without the numerical CMABs limitations~\citep{monea2025llmsincontextbanditreinforcement}. A bandit algorithm can be obtained with an LLM-based setup by appending past \textit{rounds} of interaction --- consisting of a context, action and reward tuple --- to the language model's prompt~\footnote{We will henceforth use the term \textit{prompt} to refer to any information that is used as an input to the decoder-based LLM, and exclusively use the term \textit{context} for the contextual variables in the bandit setting. This is to disambiguate from the common use of the term \textit{context} in the language modelling literature.}. The number of rounds in the prompt is limited by the model's inference capacity, a common feature of state-of-the-art LLMs. In these LLM-based CMABs, exploration is induced by randomly dropping out past rounds from the prompt and increasing the sampling temperature as demonstrated in recent works by \citet{monea2025llmsincontextbanditreinforcement} and \citet{sun2025largelanguagemodelenhancedmultiarmed}. Such methods require additional hyperparameter tuning and underutilize established CMAB literature~\citep{srinivas2009gaussian, lu2010contextual, lattimore2020bandit, Sutton1998}. Our work seeks to reconcile the differences between recent LLM-based and well-established numerical approaches. 

Crucially, using LLMs in problems where a CMAB solution is expected to perform well requires reliable uncertainty estimates that LLMs are limited in providing~\citep{hager2025uncertainty,hobelsberger2025systematic,tomov2025illusion, ulmer2025anthropomimetic}. This contrasts with the large number of numerical bandit algorithms that directly handle uncertainty like Gaussian Process Upper Confidence Bound (GP-UCB) and Thompson Sampling (TS)~\citep{lattimore2020bandit}. To address this, we turn to the concepts behind LLM Processes~\citep{requeima2024llmprocessesnumericalpredictive} where the model is queried multiple times on the same target to obtain a distribution rather than a single point prediction. 

We evaluate recent approaches to incorporating language into decision making. We include datasets of both numerical and textual features, compare LLM-based approaches to the numerical ones and propose a new competitive method that enables uncertainty estimation for bandits using LLMs. Our motivation is to combine the language understanding of LLMs with the uncertainty estimates benefits of traditional bandit algorithms to ensure more reliable and flexible decision-making in challenging settings where more complex formalisms fail and both language and numerical information are required for optimal performance.

To summarize, our contributions are:

\noindent % Prevents the entire block from being indented
\vspace{-7mm}
\begin{itemize}[leftmargin=2em, labelsep=0.5em, itemsep=0.5ex] 

    \item A \textbf{comprehensive evaluation} of LLM and non-LLM-based CMAB methods for decision-making problems where the context information contains both textual and numerical features;
    
    \item Introducing \textbf{LLM Process UCB} bandits combining the benefits of uncertainty estimation used by numerical bandits and the useful semantic priors of LLMs;

    \item Showing that \textbf{numerical bandits maintain competitive performance} when trained directly on language embeddings;

    \item Empirically demonstrating that \textbf{the dimensionality of the embeddings space directly affects exploration} and can be used to predict the performance gains resulting from using LLMs.

\end{itemize}

\vspace{-5pt}

\section{Preliminaries}
\vspace{-5pt}

In this section, we formally define the action space, different compositions of the context space and different types of reward functions. 

\subsection{Contextual Multi-Armed Bandits}
\vspace{-3pt}

\textbf{Context Space} We combine numerical and textual information. At each round $t \in [T]$, the environment presents a context $c_t = [x_t, z_t]_{d+n}$ that is a stacking of a $d$-dimensional numerical context vector $x_t \in \mathbb{R}^d$ and the $n$-dimensional textual information $z_t$ (e.g., natural language description, categorical features). Here, $n$ can be arbitrarily large as it corresponds to the size of the vocabulary in a natural language setting.

We may use the embedding function $\psi: \mathcal{Z} \rightarrow \mathbb{R}^m$ that maps textual information $z_t$ from the text space $\mathcal{Z}$ to an $m$-dimensional vector representation where $m$ is the embedding size. This would lead to $c_t = [x_t, z_t]_{d+m}$ where $m << n$. 

\textbf{Action Space} We use discrete action space $\mathcal{A}$ with $|\mathcal{A}| = K$ arms. Our \textbf{objective} is to learn a policy $\pi: \mathcal{C} \rightarrow \mathcal{A}$ that maximizes the reward over $T$ rounds $R_T= \sum_{t=1}^T r_t$. The \textbf{reward} is a mapping $R: \mathcal{C} \times \mathcal{A} \rightarrow \mathbb{R}$. For every pull $t$,  $r_{t} = f(c_t,a_t) + \epsilon_t$ where $\epsilon_t$ is sampled from a $\mathcal{N}(0,\sigma_t^2)$. We use constant $\sigma$ unless stated otherwise.

\subsection{Embeddings Models}
\vspace{-3pt}

\textbf{Matryoshka Embeddings} To use numerical bandits on traditionally language-based settings, we use the embedding function $\psi: \mathcal{Z} \rightarrow \mathbb{R}^m$ to map textual information $z_t$ into a vector representation of size $m$. Unlike standard \textbf{dense} embeddings, which require the entire vector to function correctly, \textbf{Matryoshka} embeddings~\citep{kusupati2024matryoshkarepresentationlearning} are trained on a loss function with separate terms for different sub-dimensions of the vector to increase the granularity of representation with the number of dimensions. This allows us to use just the first $k$ dimensions (where $k \ll m$) to form a smaller representation without losing the core semantic meaning. As $k$ increases, the embedding takes into account finer details of the data.

\subsection{Types of Reward Functions}
\vspace{-3pt}

To perform informative evaluations, we implement a wide range of linear and non-linear reward functions with varying dependencies on numerical and language context.

\textbf{Linearity} 
In the case of a linear reward function, the reward for a single pull $t$ is the inner product of the action-context vector $[a_t, c_t]$ and the reward weights $\theta^*$ dictating the value of each action for a given context. A function $f$ is \emph{non-linear} if there exist $x, y \in \mathbb{R}^m$ and $\alpha \in (0,1)$ such that $f(\alpha x + (1-\alpha) y) \neq \alpha f(x) + (1-\alpha) f(y)$. We outline more complex functions in the Supplementary material: piecewise linear in~\ref{app:piece-wise-linear}, non-linear in~\ref{app:non_linear_numerical} and non-linear with more components in~\ref{app:highly_non-linear}.

\textbf{LLM-as-a-Judge}
When using natural language, the reward is not a direct mapping of numerical features $x$ to a scalar $r$ since language context $z$ is also relevant. For instance, when deciding which product to purchase, consumers typically consider both numerical ratings and textual information such as reviews. For problems where the context $c$ contains both numerical features $x$ and language features $z$, we implement the reward function $f_{\text{LLM}}$ by prompting an LLM with the context $c_t$ and action $a_t$ such that $r_t = f_{\text{LLM}}(c_t, a_t)$. The output can either be a \textbf{direct scalar} representing the overall reward for the action in the given context, or a \textbf{rubric} where the LLM returns a vector of scores pertaining to a predefined rubric e.g., relevance, correctness, creativity. Implementation details in Supplementary Material~\ref{app:llm-as-a-judge}. 

\textbf{Feature Extraction}
Drawing from the established Natural Language Processing (NLP) literature, we use feature extraction methods to derive structured scores, mirroring a rubric-based evaluation. These scores are used to determine the reward without querying any LLMs. We include this type of reward function for a more challenging setting that does not use an LLM's semantic priors. We provide a more detailed description and implementation in Supplementary Material~\ref{app:feature-extraction-based-reward}.

\vspace{-10pt}

\section{Related Work}
\label{sec:related_work}
\vspace{-7pt}

In this section we provide an overview of related work in the space of bandits and LLMs for sequential decision making. A more extended version can be found in~\autoref{app:extended_related_work}.

\textbf{Prompt-based Learning} LLMs have been shown to increase performance when prompted with examples of successful input-output pairs, a phenomenon commonly referred to as \textit{in-context} learning~\citep{brown2020language}; in this work we refer to it as \textit{in-prompt} learning to avoid confounding terminology with preceding contextual bandits literature. \citet{agarwal2024manyshotincontextlearning} demonstrated that performance scales with the number of examples in the prompt, which is limited by the prompt's maximum length and is sensitive to its ordering~\citep{lu2022fantasticallyorderedpromptsthem} and formatting~\citep{ahmed2025intent, sgouritsa2024promptingstrategiesenablinglarge}. \citet{wenliang2025prompting} further show that intuitive few-shot prompting is suboptimal and that understanding prompting strategies remains an open area of scientific inquiry.

\citet{genewein2025understanding} also confirm that using non-intuitive, real-valued vectors outside the token alphabet increases effectiveness and point to limitations that can only be overcome by further training. Similarly, \citet{mukherjee2025pretrainingdecisiontransformersreward} and~\citet{kirsch2024generalpurposeincontextlearningmetalearning} have shown that causal transformers can be meta-trained to act as in-context learners, which is beyond the scope of this work. 

\textbf{LLM-based Stochastic Processes} Recent work expands the use of in‑context learning and prompt engineering to approximate text‑conditioned stochastic processes. \citet{requeima2024llmprocessesnumericalpredictive} demonstrate that with carefully formatted prompts, LLMs can be used to obtain reliable point predictions and uncertainty estimates. The \textit{Context is Key} benchmark by~\citet{williams2025contextkeybenchmarkforecasting} provides comprehensive comparisons demonstrating that conditioning on language and past observations improves prediction accuracy under rare events without additional covariates.

\textbf{Combining Bandits and LLMs} The ``hybrid'' approach by \citet{baheri2023llmsaugmentedcontextualbandit} uses the LLM layers closer to the final output to encode language in a numerical vector to be used in traditional CMAB methods. Although promising, their method's evaluation is limited to a synthetic dataset and relies on having access to the LLM weights. \citet{chen2025efficientsequentialdecisionmaking}'s method relies more heavily on LLM queries as warm-starting guidance, then gradually anneals to standard bandit algorithms.

\citet{monea2025llmsincontextbanditreinforcement} study the ability of LLMs to operate as contextual bandits by implementing several algorithms and prompt composition strategies to balance exploration tradeoffs. They achieve competitive results on more realistic classification problems in financial and clinical domains. \citet{song2025investigatingrelationshipphysicalactivity} also study a ``hybrid approach'' where the CMAB selects the actions that trigger the LLM generation on which the reward is determined.

\citet{sun2025largelanguagemodelenhancedmultiarmed} use the LLM as a reward predictor within classical bandit frameworks. Their LLM-augmented Thompson sampling (TS-LLM) and Regression-Oracle (RO-LLM) algorithms leverage LLM's ``in-context'' or ``in-prompt'' learning capabilities to predict the reward for each arm. This approach is most similar to our work and demonstrates particular effectiveness in challenging scenarios where arms lack semantic meaning. 

\textbf{LLM Routing} A growing literature routes or selects among models to control inference cost, dispatching easier queries to cheaper models. \citet{fernandez2025radar} frame the joint choice of model size and reasoning budget as an optimization over estimates of query difficulty and model-budget abilities. This work shares our motivation of spending LLM capacity only where it helps, but the role of the bandit differs: instead of allocating generation effort \textit{across} LLMs, we explore CMABs on text embeddings as a substitute \textit{for} LLM reasoning.

\section{Problem Statement}
\label{sec:problem_statement}

We set out to examine CMABs involving textual, numerical, or combined contexts. We illustrate the variable role of LLMs within the decision pipeline in \autoref{fig:bandits}. We focus on two core questions:
% For completeness, we also test purely numerical bandits in scenarios where the optimal action depends on both language and numerical data. 

\begin{figure}[t]
    \centering
    \includegraphics[width=0.8\textwidth]{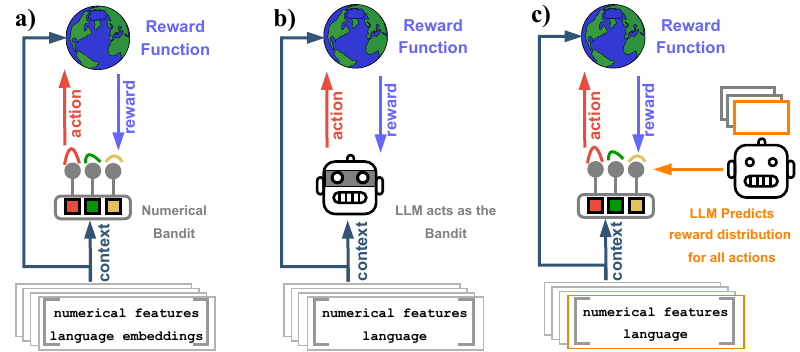}
    \caption{A taxonomy of the bandit architectures used in this work. a) and c) illustrate the case where we perform reward function regression for each arm. Alternatively, in b) the LLM makes the decision itself without modeling the expected reward of each arm. \vspace{-7pt}}
    \label{fig:bandits}
\end{figure}

\begin{center}
\begin{minipage}{\textwidth} % Adjust 0.9 to make it wider or narrower
    \textbf{\textcolor{purple}{Q1}} \textit{When considering contextual bandit problems where the context contains both numerical and language information, when are LLM-based bandits better than traditional contextual bandits?}
    
    \vspace{3mm}
    
    \textbf{\textcolor{cyan}{Q2}} \textit{When building natural language-guided contextual bandits, are LLM-Process based modeling approaches better than in-context reinforcement learning using LLMs?}
\end{minipage}

\end{center}

In \autoref{fig:bandits}, it should be noted that approach \textbf{b)} has been abundantly explored in the literature~\citep{monea2025llmsincontextbanditreinforcement, song2025investigatingrelationshipphysicalactivity,baheri2023llmsaugmentedcontextualbandit}, whereas a simpler version of \textbf{c)} is used by~\citet{sun2025largelanguagemodelenhancedmultiarmed} where the LLM is used to provide a single reward estimate rather than a distribution. We can relate the research questions to the different setups in~\autoref{fig:bandits}. 

\textbf{\textcolor{purple}{Q1}} can be thought of as: \textit{What is better, approach a) or one of b) or c) from~\autoref{fig:bandits}?} \textbf{\textcolor{cyan}{Q2}} can be thought of as: \textit{Which one is better, approach b) or c) from~\autoref{fig:bandits}?}

\textbf{Numerical CMABs} Corresponds to \textbf{a} in~\autoref{fig:bandits}. \textbf{LinUCB} maintains a linear model for each arm, updating parameters with observed context and reward. \textbf{Thompson Sampling} maintains a probabilistic model for each arm, sampling model parameters from their posterior distribution. \textbf{Contextual Gaussian Process  (CGP-UCB)} uses a Gaussian Process (GP) to model the reward as a function of both context and action. It selects arms using a UCB criterion based on the GP's predicted mean and standard deviation. For the GP we use a Radial Basis Function (RBF) kernel.  

\textbf{LLM Bandits} Corresponds to \textbf{b} in~\autoref{fig:bandits},  uses the action, context, reward history to return the next action. Directly prompts the LLM to do in-context reinforcement learning (ICRL).

\textbf{LLM Process} Corresponds to \textbf{c},  makes the mean and standard deviation prediction conditioned on language and numerical context for every action. For each arm $a \in \mathcal{A}$, we query the LLM $q$ times with a fixed prompt. The $q$ responses are used to estimate mean $\hat\mu_a$ and standard deviation $\hat\sigma_a$ of the reward for each arm. The next arm selection is based on the score function $S(a) = \hat\mu_a + \beta\hat\sigma_a$ with the $\beta$ hyperparameter \footnote{$\beta$ was not tuned in this work. We only used the default $\beta=1$}. Both epistemic and sampling variance are included in $\sigma$. We induce randomness by using a softmax temperature of $0.6$ for the LLM. To the best of our knowledge, this work is the first one to combine the LLM Process and contextual bandits. We outline our method in Algorithm~\ref{alg:llm_process_bandit}.
\textbf{LLM Process Joint} is a `joint' variant of our method, distinct in the computation of line~\ref{algline:arm_loop} in Algorithm~\ref{alg:llm_process_bandit}, which is done by asking for the values of all the arms in the \textit{same prompt}. This is done the same number of times as the \textbf{LLM Process} does for each individual arm. Both with the \textbf{LLM Process} and the \textbf{LLM Process Joint}, the $q$ identical queries are done in parallel through \texttt{vLLM}'s optimized inference~\citep{kwon2023efficient}.

\begin{algorithm}[H]
\footnotesize
\caption{LLM Process Bandit (LLMP-UCB)}
\label{alg:llm_process_bandit}
\begin{algorithmic}[1]
\State \textbf{Inputs:} Total steps $T$; Environment $E$; Action set $\mathcal{A}$
\State $c_{0}, \_ \leftarrow E.\text{reset}()$ \hfill{\color{gray}{\emph{Initialize environment context}}}
\State $H \leftarrow \emptyset$
\For{$t=1$ to $T$}
    \For{each arm $a \in \mathcal{A}$} \hfill{\color{gray}{\emph{LLM Process evaluation}}} \label{algline:arm_loop}
        \State $\mathcal{R}_{a} \leftarrow \emptyset$
        \For{$i=1$ to $q$}
            \State $\hat{r}_{i} \leftarrow \text{LLM}(c_{t-1}, a, \text{temp}=0.6)$ \hfill{\color{gray}{\emph{Sample predicted reward}}}
            \State $\mathcal{R}_{a} \leftarrow \mathcal{R}_{a} \cup \{\hat{r}_{i}\}$
        \EndFor
        \State $\hat{\mu}_a \leftarrow \text{mean}(\mathcal{R}_{a})$
        \State $\hat{\sigma}_a \leftarrow \text{std}(\mathcal{R}_{a})$ \hfill{\color{gray}{\emph{Includes epistemic and sampling variance}}}
        \State $S(a) \leftarrow \hat{\mu}_a + \beta\hat{\sigma}_a$ \hfill{\color{gray}{\emph{Compute score function}}}
    \EndFor
    
    \State $a_t  \leftarrow \text{argmax}_{a \in \mathcal{A}} S(a)$ \hfill{\color{gray}{\emph{Select arm with highest score}}}
    
    \State $c_t, r_t, \text{done} \leftarrow E.\text{step}(a_t)$   \hfill{\color{gray}{\emph{E returns reward $r$ and the next context $c_t$}}}\label{algline:env_step}
    
    \State $\mathcal{B}.\text{update}(c_{t-1}, a_t, r_t, c_t)$ \hfill{\color{gray}{\emph{Update bandit knowledge}}}
    
    \State $H \leftarrow H \cup \{(r_t, \text{Regret}, a_t, c_{t-1})\}$ 
    
    \If{\text{done}}
        \State $c_t, \_ \leftarrow E.\text{reset}()$ \hfill{\color{gray}{\emph{Reset if episode done}}}
    \EndIf
\EndFor
\State \textbf{Output:} History $H$
\end{algorithmic}
\end{algorithm}

\vspace{-5pt}
Note that the optimal reward $r^*$ is only available in the case of a synthetic dataset and is only used to track absolute regret and not used by the bandit.

\vspace{-10pt}
\section{Experimental Setup}
\vspace{-5pt}

\textbf{Synthetic Dataset} We test all the bandit architectures on a synthetically generated movie recommendation dataset. Dataset generation details can be found in Supplementary Materials~\ref{app:dataset_gen}. We implement this in OpenAI's Gym environment format~\citep{openai-gym}. We test the bandits on a wide range of reward functions:  a linear reward function that is a simple lookup table of the preferred movies (\textbf{fnum\_lin}), number-theoretic piecewise linear constraints \textbf{nonlin1} (\ref{app:piece-wise-linear}), polynomial cubic or quadratic dependencies \textbf{fnum\_nonlin} (\ref{app:non_linear_numerical}) to highly non-linear transcendental functions containing sinusoids and exponentials \textbf{nonlin2} (\ref{app:highly_non-linear}), alongside semantic objectives derived through feature extraction \textbf{fextract} (\ref{app:feature-extraction-based-reward}) or direct LLM-as-a-judge \textbf{fLLM} (\ref{app:llm-as-a-judge}). 

\textbf{Language Classification Dataset} We use intent classification datasets where the bandit has to learn to classify incoming data points from past observations, receiving only a binary reward signal based on the prediction's accuracy. \textit{Banking77}~\citep{casanueva2020efficient} is a customer-complaint intent classification dataset used in commercial banking, with 77 different intent labels. The \textit{TREC} dataset~\citep{voorhees2000overview} comes in two variants, \textit{Coarse} and \textit{Fine}, and was originally designed to benchmark information retrieval from a large corpus. \textit{TREC Coarse}~\citep{li2002learning} has 6 high-level labels (e.g., ``human'', ``location''), whereas \textit{TREC Fine}~\citep{hovy2001toward} refines these into 50 finer-grained labels (e.g., ``human-being'' or ``group'' under ``human'', and ``city'' or ``country'' under ``location''). For these classification datasets, the action space consists of the labels, and the context is each customer complaint in the case of \textit{Banking77}, or the query to be answered in the case of \textit{TREC}. We list the embedding models used in Supplementary Material~\ref{app:embeddings_model_assets}.

\textbf{Baselines} In addition to the algorithms in~\autoref{sec:problem_statement}, we implement the \textbf{Regression Oracle}~\citep{foster2018practical} where we take the best results from the hyperparameter sweep ranges recommended by \citet{sun2025largelanguagemodelenhancedmultiarmed}. We include the \textbf{LLM Thompson Sampling Bandit} by~\citet{sun2025largelanguagemodelenhancedmultiarmed} as it corresponds to a single-query instance of the \textbf{LLM Process Bandit} (Algorithm~\ref{alg:llm_process_bandit}). We test the synthetic dataset in an RL-Gym environment and the language-only classification datasets on top of the LLM Bandit baselines implemented by~\citet{monea2025llmsincontextbanditreinforcement}, using the same initial prompt and task description as their implementation for a fair evaluation. \textbf{Zero-Shot} corresponds to the control of asking the LLM to choose an arm without any past rounds in the prompt and \textbf{Random} is uniformly sampling an arm with replacement every round. We validate our numerical bandit implementations in Supplementary Materials~\ref{app:numerical_baselines}. We use \texttt{Llama3.1}~\citep{grattafiori2024llama3herdmodels} and \texttt{Qwen2.5}~\citep{qwen2025qwen25technicalreport} as our LLMs in line with~\citet{monea2025llmsincontextbanditreinforcement}.

\textbf{Our method} We test the two variants of our proposed method, \textbf{LLMP Bandit} and \textbf{LLMP-Joint Bandit}, as detailed in Algorithm~\ref{alg:llm_process_bandit} and in Section \ref{sec:problem_statement}. 

\vspace{-5pt}

\section{Results}
\label{sec:res}
\vspace{-5pt}
This section contains the result of our evaluation and the performance of our proposed method.
\vspace{-5pt}

\subsection{Synthetic Dataset}
\vspace{-5pt}

\autoref{tab:regret_bandits} shows the cumulative regret and standard deviation across the synthetic benchmarks, demonstrating that our proposed \textbf{LLMP-Joint} and \textbf{LLMP-Bandit} architectures achieve state-of-the-art performance in semantically rich and complex non-linear environments. Specifically, \textbf{LLMP-Joint} achieves the best mean regret on the language-dependent tasks, achieving the lowest regret on \textbf{fextract} ($2.6 \pm 0.6$) and \textbf{fLLM} ($35.6 \pm 6.9$), suggesting that joint estimation of arm values effectively captures the nuance of language-based rewards where independent priors fail. In contrast, for the strictly linear numerical setting (\textbf{fnum\_lin}), the traditional \textbf{LinUCB} algorithm outperforms LLM-based approaches ($36.6 \pm 9.4$), indicating that expensive semantic reasoning is redundant for linear numerical reward landscapes. However, as the numerical relationships become highly non-linear (\textbf{nonlin1}, \textbf{nonlin2}), the calibrated uncertainty of our \textbf{LLMP} variants outperforms both the standard \textbf{LLM-Bandit} and \textbf{CGP-UCB} by linking semantic understanding and numerical optimization.

\begin{table}[h]
\caption{Cumulative regret and standard deviation across 5 seeds with different reward functions.}
\scriptsize
\centering
\begin{tabular}{lcccccc}
\toprule
\textbf{Algorithm} & \textbf{fextract} & \textbf{fLLM} & \textbf{fnum\_lin} & \textbf{fnum\_nonlin} & \textbf{nonlin1} & \textbf{nonlin2} \\
\midrule
CGP-UCB & $8.2 \pm 2.2$ & $65.60 \pm 8.04$ & $199.80 \pm 25.1$ & $444.9 \pm 45.6$ & $315.0 \pm 28.2$ & $235.3 \pm 27.0$ \\
LinUCB & $10.2 \pm 1.9$ & $43.2 \pm 7.3$ & \textbf{36.6 $\pm$ 9.4} & $351.9 \pm 45.5$ & $181.6 \pm 26.3$ & \textbf{23.0 $\pm$ 7.2} \\
LLM-Bandit & $11.5 \pm 1.0$ & $36.0 \pm 6.9$ & $190.0 \pm 16.3$ & \textbf{302.7 $\pm$ 42.0} & $346.6 \pm 28.7$ & $43.7 \pm 10.6$ \\
LLMP-Bandit & $17.2 \pm 2.9$ & $40.00 \pm 7.1$ & \textbf{54.2 $\pm$ 11.4} & $378.6 \pm 36.9$ & \textbf{150.0} $\pm$ \textbf{22.4} & \textbf{23.2 $\pm$ 9.6} \\
Random & $27.6 \pm 3.7$ & $130.4 \pm 8.0$ & $256.2 \pm 25.2$ & $568.2 \pm 48.2$ & $297.2 \pm 26.0$ & $297.5 \pm 27.8$ \\
Thompson & $14.9 \pm 2.4$ & $54.6 \pm 7.8$ & $54.0 \pm 12.4$ & $317.1 \pm 43.1$ & \textbf{152.8} $\pm$ \textbf{19.3} & $72.9 \pm 14.7$ \\
ZeroShot & $12.7 \pm 0.9$ & $180.6 \pm 4.0$ & $220.0 \pm 14.5$ & $885.2 \pm 30.5 $ & $265.2 \pm 14.8$ & $258.6 \pm 16.6$ \\
LLMP-Joint &\textbf{ 2.6 $\pm$ 0.6} & \textbf{35.6 $\pm$ 6.9} & $65.7 \pm 14.1$ & $374.2 \pm 43.8$ & \textbf{150.0 $\pm$ 23.2} & $103.7 \pm 20.6$ \\
\bottomrule
\end{tabular}
\label{tab:regret_bandits}
\vspace{-10pt}
\end{table}

\begin{figure}[h]
    \centering
    \includegraphics[width=0.9\textwidth]{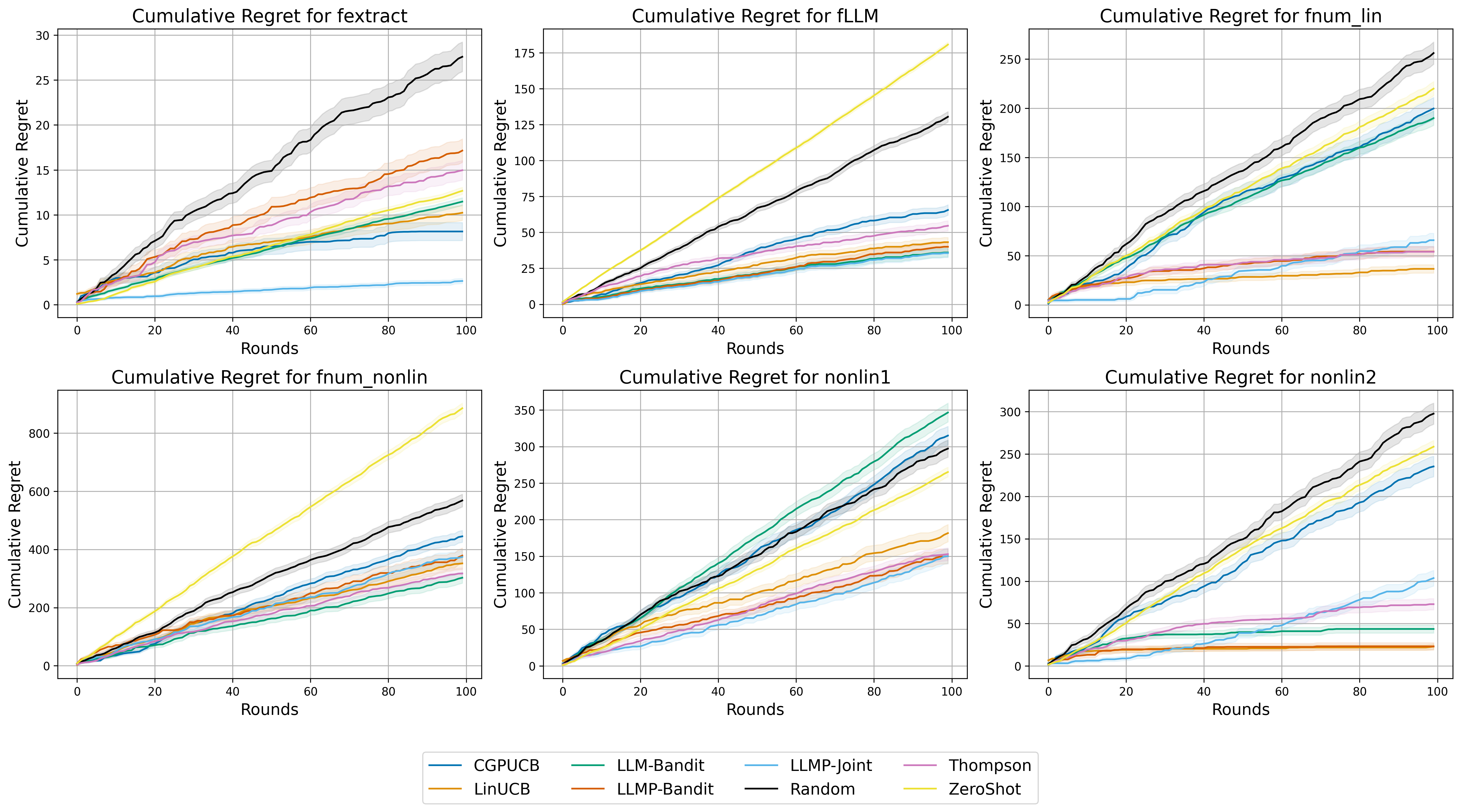}
    \caption{Full experimental results of the different algorithms and reward functions\vspace{-7pt}}
    \label{fig:movie_dataset_results_full}
\end{figure}

\autoref{fig:movie_dataset_results_full} shows the temporal evolution of \textit{cumulative} regret, revealing distinct learning dynamics. In the \textbf{fextract} reward function with the extracted features, \textbf{LLMP-Joint} demonstrates superior sample efficiency, with its regret curve flattening significantly earlier than other baselines. We therefore posit that modeling the joint distribution of arm values is critical for capturing correlated semantic features. Conversely, the strictly linear \textbf{fnum\_lin} setting exposes the drawback of this approach: here, the specialized \textbf{LinUCB} algorithm converges fastest, whereas LLM-based methods incur higher regret by attempting to find complex patterns in a simple linear surface. However, the plots also highlight the sensitivity of numerical methods in complex settings as \textbf{LinUCB} fails to learn in non-linear environments like \textbf{nonlin1} (exhibiting linear regret growth) while our LLM-based agents maintain sub-linear trajectories where traditional linear assumptions break down. In all cases, one of the two variants we propose, \textbf{LLMP-Bandit} or \textbf{LLMP-Joint}, matches or surpasses the traditional LLM Bandits where the LLM chooses the arm directly instead of only being used for value estimation. 

\vspace{-7pt}

\subsection{Language Classification Datasets}

\begin{figure}[h!]
    \centering
    \includegraphics[width=\textwidth]{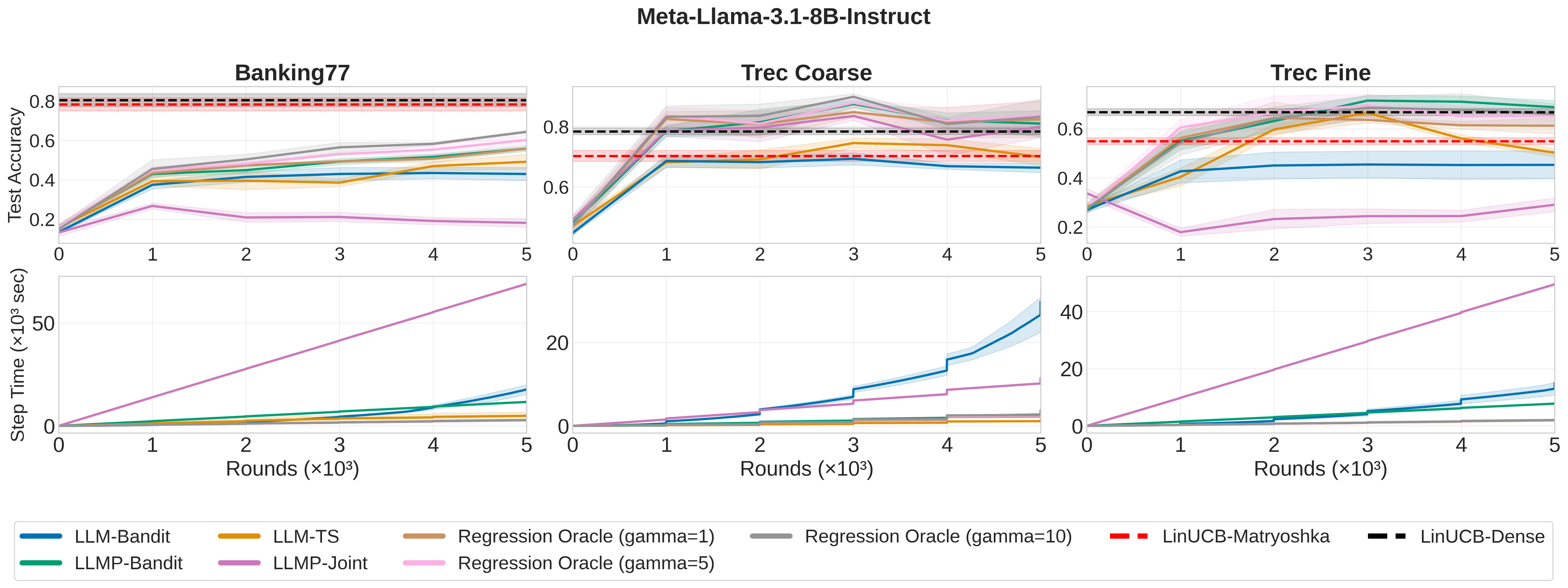}
    \caption{\texttt{Llama3.1-8B} Results for all the methods with the numerical baselines as horizontal lines.\vspace{-9pt}}
    \label{fig:Llama_res_all}
\end{figure}

\begin{figure}[h!]
    \centering
    \includegraphics[width=\textwidth]{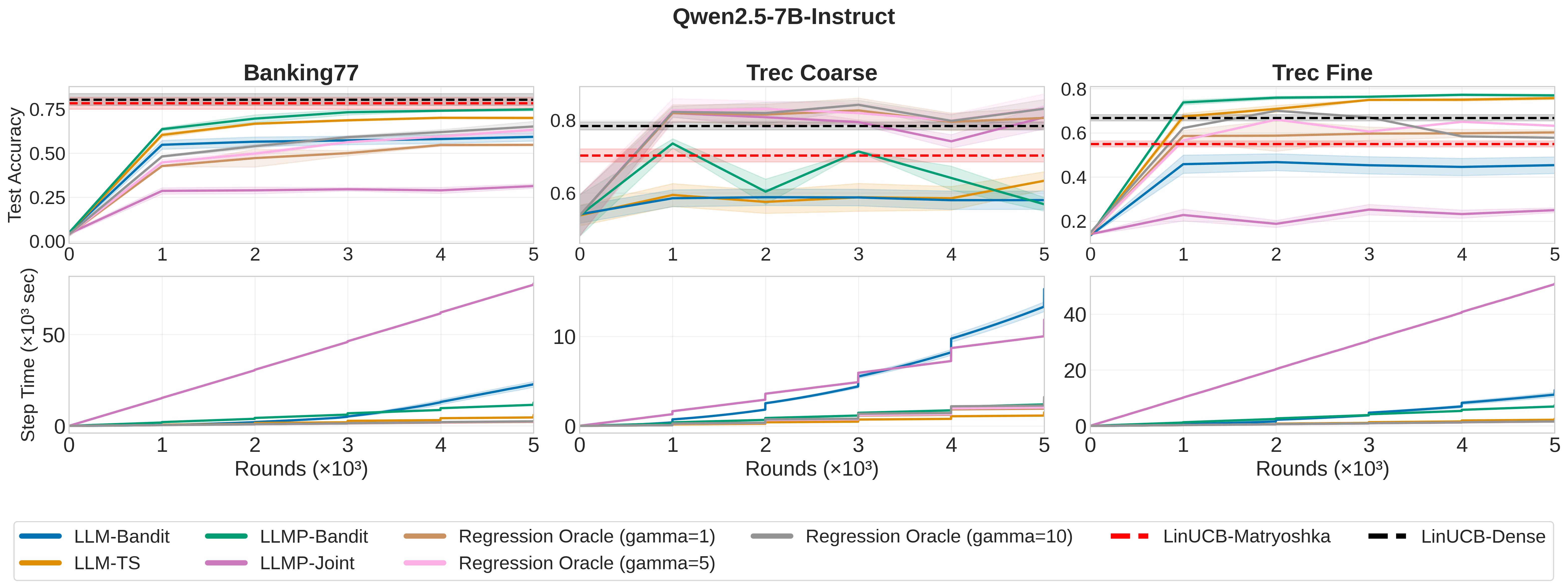}
    \caption{\texttt{Qwen2.5-7B} Results for all the methods with the numerical baselines as horizontal lines. \vspace{-5pt}}
    \label{fig:Qwen_res_all}
\end{figure}

\autoref{fig:Llama_res_all} and \autoref{fig:Qwen_res_all} show a clear tradeoff between the complexity of the action space and the performance of our proposed methods. \textbf{LLMP-Joint} performs poorly on \textbf{Banking77}, which contains 77 classes with high semantic similarity (e.g., distinguishing between \texttt{declined\_card\_payment} and \texttt{declined\_cash\_withdrawal}), dropping well below the accuracy of the \textbf{LLMP-Bandit} and standard \textbf{LLM-Bandit}. Jointly evaluating a large number of subtly different options does not make the best use of the limited prompt length. However, on \textbf{TREC Coarse} (containing only 6 distinct categories), \textbf{LLMP-Joint} performs competitively. This indicates that the joint method is effective for small action spaces but underperforms when the arms in the prompt are semantically similar.

In Figure~\ref{fig:Llama_res_all} and~\ref{fig:Qwen_res_all}, the horizontal dashed lines show that standard numerical bandits operating on static \textbf{Dense} or \textbf{Matryoshka} embeddings often match or outperform the complex online LLM agents in tasks where the arms are semantically similar. Additionally, the step time plots (bottom rows) reveal that \textbf{LLMP-Joint} is the most computationally expensive approach. Due to having to generate a score for every arm at every step, \textbf{LLMP-Joint} incurs a much higher inference cost, making it less suitable for high-dimensional problems like \textbf{Banking77} despite its theoretical advantages with smaller action spaces.

\section{Discussion}
\label{sec:discussion}

In this section, we interpret the results to explicitly answer the research questions posed in \autoref{sec:problem_statement}. 

\textcolor{purple}{\textbf{Q1} on comparing LLM-based and traditional bandits}:  Our results indicate that the superiority of LLM-based approaches is more closely tied to the complexity of reward functions rather than the mere presence of natural language in the context. In environments with complex, non-linear \textit{numerical} reward functions like \textbf{nonlin2} and \textbf{nonlin1}, LLM-based bandit architectures perform best alongside at least one of the numerical CMABs. In any setting where the reward function is determined either directly from language by an LLM like \textbf{fLLM} or by language feature extraction like \textbf{fextract}, LLM Process bandits outperform the rest. However, for standard classification tasks like \textbf{Banking77}, we find that traditional numerical bandits are surprisingly effective.

\textcolor{cyan}{\textbf{Q2} on LLM Processes}: We observe that explicitly modeling reward distributions (\textbf{LLMP}) generally outperforms direct in-context learning by providing a principled mechanism for uncertainty. However, while the \textbf{LLMP-Joint} variant excels in low-dimensional action spaces by capturing relative values, it degrades in high-dimensional settings like \textbf{Banking77} due to context window constraints and semantic similarity of the arms. While process-based uncertainty is effective, independent estimation or efficient numerical methods are preferable for high-dimensional action spaces.

\begin{figure}[h]
    \centering
    \includegraphics[width=0.95\textwidth]{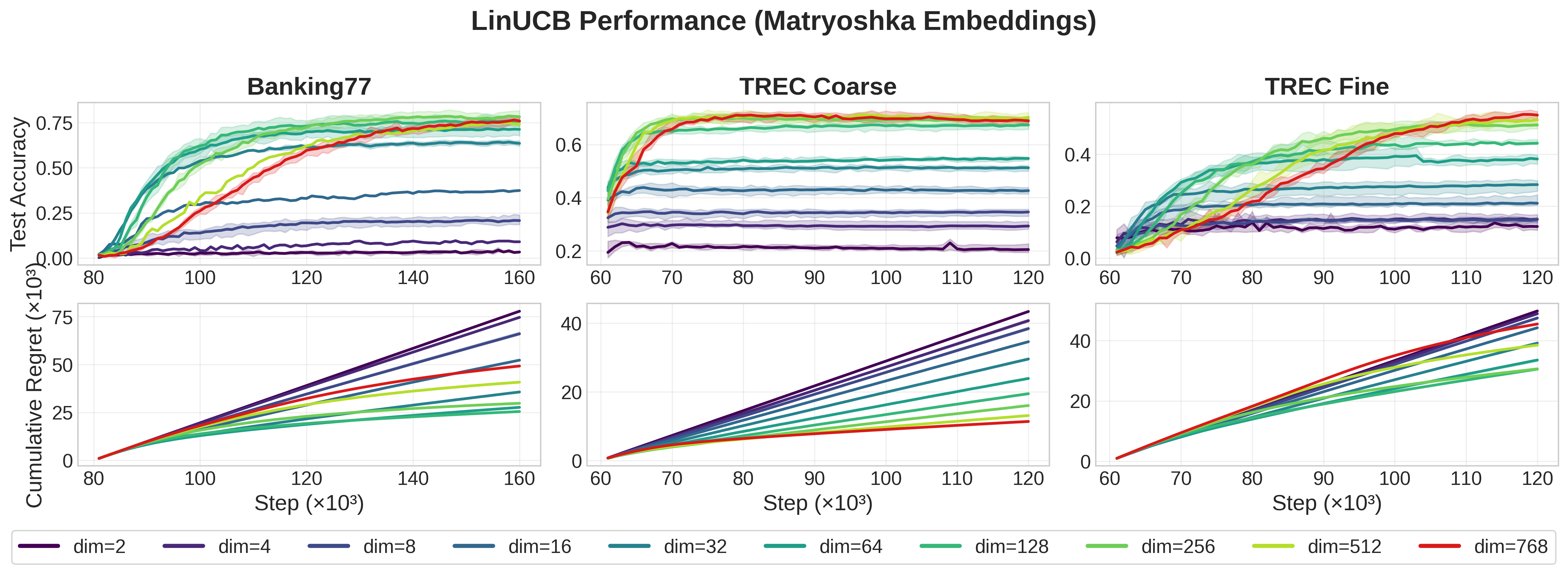}
    \caption{LinUCB results for all dimensions when using Matryoshka representation. \vspace{-7pt}}
    \label{fig:full_matryoshka_res}
\end{figure}

\begin{figure}[h]
    \centering
    \includegraphics[width=0.95\textwidth]{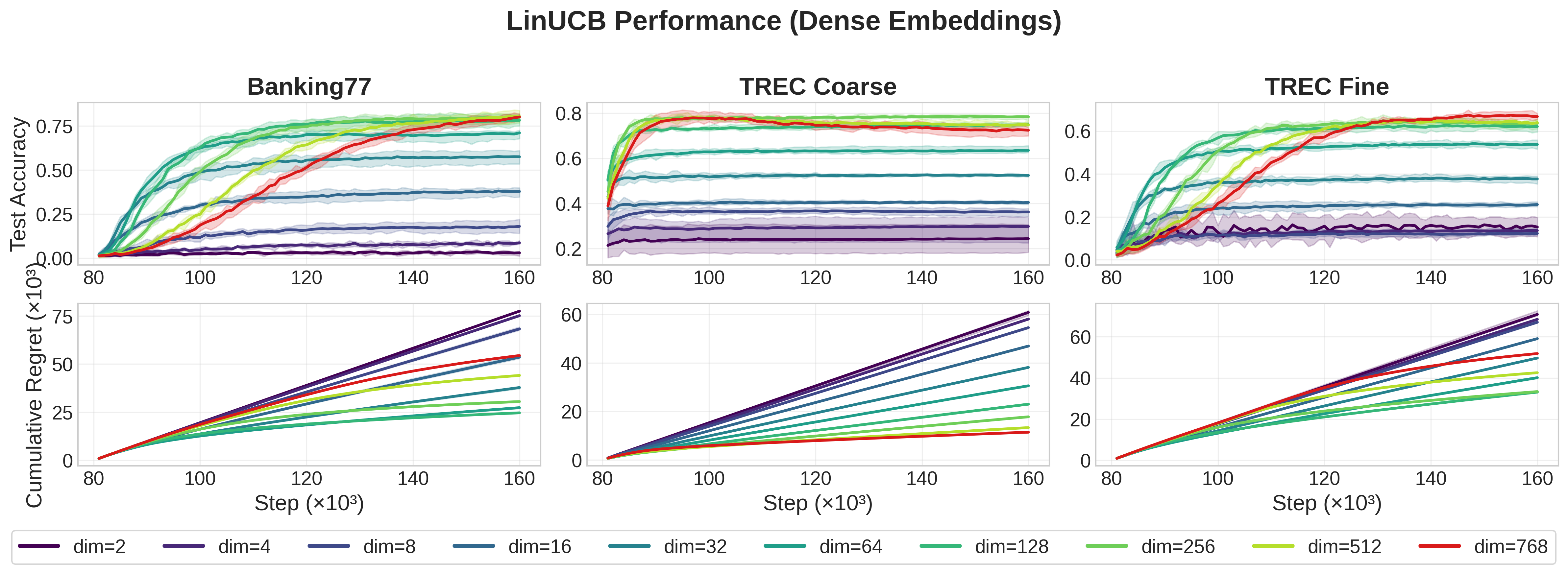}
    \caption{LinUCB results for all dimensions when using traditional dense representation.}
    \label{fig:full_dense_res}
\vspace{-10pt}
\end{figure}

To investigate the competitive performance of numerical CMABs, we analyze how the language embeddings affect learning. \autoref{fig:full_matryoshka_res} reveals that the number of dimensions acts as a direct control for the exploration-exploitation tradeoff. When using fewer dimensions (e.g., 2 to 16), the \textbf{LinUCB} agent learns very fast because the search space is small, but it plateaus at a low accuracy. In contrast, using more dimensions (up to 768) allows the agent to reach a higher final accuracy at a higher sampling complexity. This allows direct control of the exploration-exploitation tradeoff simply by changing the input embedding size. These results hold for dense embeddings too (\autoref{fig:full_dense_res}), with \textbf{Matryoshka} having more distinct final performances in the lower dimensions.

% \vspace{-10pt}

\begin{figure}[h]
    \centering
    \includegraphics[width=1.0\textwidth]{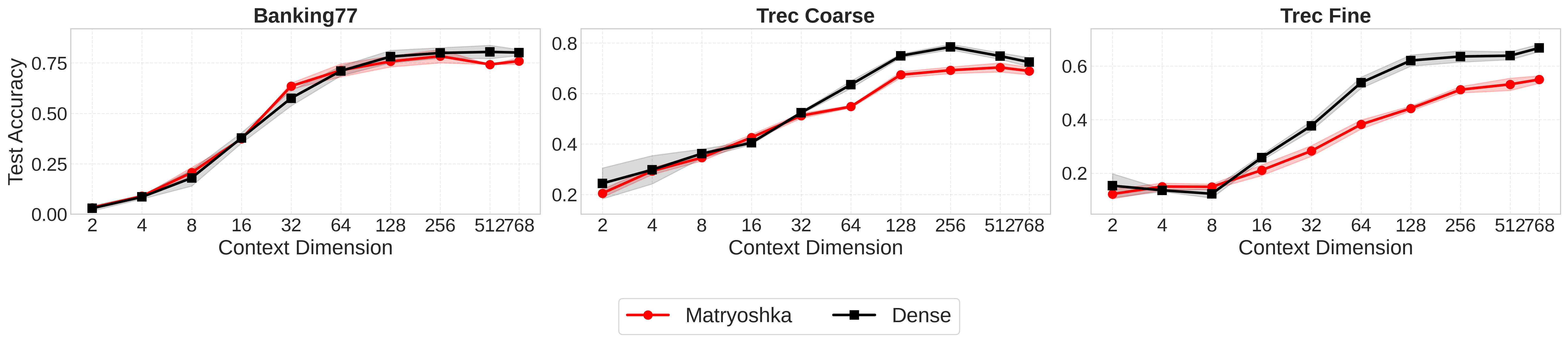}
    \caption{LinUCB final accuracies across dimensions for all the types of embeddings used.\vspace{-7pt}}
    \label{fig:Dense_and_Matryoshka_comparison}
\end{figure}

\autoref{fig:Dense_and_Matryoshka_comparison} directly compares the best performance of the two ways of encoding textual information. \textbf{Dense} embeddings perform better at medium-sized dimensions, allowing the linear bandit to distinguish between classes. When using the full embedding size, both types allow the simple \textbf{LinUCB} algorithm to effectively separate semantically close arms. This suggests that a simple numerical bandit operating on high-dimensional embeddings can be sufficient on language tasks.

\vspace{-8pt}

\subsection{Practical Diagnostic}
\label{sec:diagnostic}

Our results yield a two-step rule for choosing between an LLM-based and a
numerical bandit. The question is geometric: can the reward be recovered from
proximity in embedding space, or does it require deeper understanding of the text?
    
     \begin{tcolorbox}[
      colback=gray!10,
      colframe=gray,
      coltitle=white,
      fonttitle=\bfseries,
      title=A Two-Step Diagnostic
    ]
    
    \textbf{Step 1 (free, \emph{a priori} intuition):} If the reward needs \emph{semantic
    judgment} or in-context reasoning beyond embedding similarity, an LLM is strongly advised. If the reward depends only
    on numerical or structural relationships, \emph{even non-linear ones},
    a numerical bandit suffices. In Table~\ref{tab:regret_bandits} this separates
   \textbf{fextract} and \textbf{fLLM} (LLMP wins) from the numerical rewards cases where
    LinUCB is competitive or the best.
    
    \vspace{7pt}
    
    \textbf{Step 2 (in case Step 1 results are inconclusive):} Sweep LinUCB over embedding
    dimensions (Figs.~\ref{fig:full_matryoshka_res},~\ref{fig:full_dense_res}), which is orders of
    magnitude cheaper than an LLM agent in terms of both time and deployment logistics (step times, Figs.~\ref{fig:Llama_res_all},~\ref{fig:Qwen_res_all}).
    If accuracy plateaus near the LLM baseline at low-to-medium dimensions, the LLM
    is not justified. 
    %Otherwise, the LLM's semantic priors are needed.
    \end{tcolorbox}

Note that \textbf{Step 1} is a heuristic prior rather than a formal test. \textbf{Step 2} is therefore the fallback whenever this judgment is uncertain, giving an empirical decision that does not depend on intuition.

A notable observation is that the gap between the Matryoshka and dense embedding performance on low-to-medium dimensions could further indicate that a
numerical bandit suffices (e.g. Banking77, Fig.~\ref{fig:Dense_and_Matryoshka_comparison}). We
leave broader validation of this latter observation to future work.

\section{Conclusion}

Our work demonstrates that deploying expensive LLMs is not strictly necessary for effective decision-making in contextual bandit settings where both numerical and text-based features are present. Moreover, drawing on recent techniques for uncertainty quantification in LLMs, we propose LLMP-UCB, a novel bandit algorithm to address this setting. By investigating the geometry of the problem, we show that lightweight numerical bandits can often match or exceed the performance of complex reasoning agents. We put forward an embedding-based assessment that allows practitioners to select the most efficient solution, deploying deep LLM uncertainty estimation only when required by the complexity of the reward function. We distill this into a two-step diagnostic in Section~\ref{sec:diagnostic}. Future work requires detailed analysis across embedding models and additional experiments to disentangle the quality of the embedding model from the intrinsic difficulty of the problem. We further see a fully formal characterization of when embeddings are sufficient, for example, a PAC-style bound on the embeddings' ability to recover different types of reward functions, as an important open problem. We view our empirical diagnostic as a necessary first step toward this goal.

\section*{Disclaimer}
This paper was prepared for informational purposes in part by the Quantitative Trading \& Research Group of JPMorganChase \& Co. This paper is not a product of the Research Department of JPMorganChase \& Co. or its affiliates. Neither JPMorganChase \& Co. nor any of its affiliates makes any explicit or implied representation or warranty and none of them accept any liability in connection with this paper, including, without limitation, with respect to the completeness, accuracy, or reliability of the information contained herein and the potential legal, compliance, tax, or accounting effects thereof. This document is not intended as investment research or investment advice, or as a recommendation, offer, or solicitation for the purchase or sale of any security, financial instrument, financial product or service, or to be used in any way for evaluating the merits of participating in any transaction, and shall not constitute a solicitation under any jurisdiction or to any person, if such solicitation under such jurisdiction or to such person would be unlawful.

\newpage
\bibliography{bib/main}
\bibliographystyle{rlj}

\beginSupplementaryMaterials

Our supplementary materials sections are structured as follows:

\begin{itemize} 

\item \cref{app:extended_related_work} provides an extended literature review.

\item \cref{app:reward_function_ontology} details the mathematical definitions and properties of the linear, non-linear, and LLM-based reward functions used in our synthetic experiments.

\item \cref{app:numerical_baselines} validates the implementation of our numerical bandit baselines by testing on linear and non-linear reward surfaces.

\item \cref{app:dataset_gen} outlines the generation methodology and data schema for the synthetic movie recommendation dataset created via LLM prompting.

\item \cref{app:embeddings_model_assets} lists the licensing information for the specific embedding models used for numerical representations and the public benchmarks used for the language classification problems (Banking77 and TREC).

\end{itemize}

\newpage

\appendix
\section{Extended Related Work}
\label{app:extended_related_work}

\subsection{In-Context Reinforcement Learning} 
Reinforcement learning (RL)~\citep{Sutton1998} is a more challenging setting where the training distribution changes as the agent interacts with the environment, which is defined by a Markov Decision Process (MDP) with a transition and reward function~\citep{Sutton1998}. The agent's ability to perform well on out-of-distribution tasks without any additional training or fine-tuning is commonly referred to as In-Context Reinforcement Learning (ICRL)~\citep{moeini2025surveyincontextreinforcementlearning}, a term first coined by~\citep{laskin2022incontextreinforcementlearningalgorithm} in their Algorithmic Distillation (AD) work. In this approach, causal transformers are trained on multi-task trajectories to predict the optimal action without being conditioned on any optimal reward signal, unlike Decision Transformer~\citep{chen2021decisiontransformerreinforcementlearning} where the agent is conditioned on the optimal cumulative reward discounted by the rewards received during inference. This inference-time phenomenon is largely attributed to the pretrained network implicitly implementing an RL algorithm in the forward pass. 

\citet{codaforno2023metaincontextlearninglargelanguage} show that LLMs can meta-learn from multiple bandit problems appended to the prompt. \citet{song2025rewardenoughllmsincontext} focus exclusively on tasks requiring general knowledge that can only be introduced via LLMs. Their algorithm appends previous action and reward pairs to the prompt which then explicitly asks for a choice between exploration and exploitation. \citet{schmied2025llms} test their LLM Agents on multi-armed bandit problems and simple games and delineate the main pathologies as being sub-optimal exploration due to a frequency and greediness bias and the inability to effectively act on knowledge already present in the model.

\subsection{LLMs for Numerical Predictions} 

Our LLMP-UCB algorithm uses the LLM Process methods from \citet{requeima2024llmprocessesnumericalpredictive} and the desiderata from the \citet{williams2025contextkeybenchmarkforecasting}'s \textit{Context is Key} benchmark where time series forecasting pairs numerical data with textual information to improve accuracy. We note that earlier work by~\citet{gruver2023large} also finds that LLMs, specifically earlier ones like GPT-3~\citep{brown2020language} and LLaMa-2~\citep{touvron2023llama}, can be used as \textit{zero-shot} time series forecasters and match the performance of trained numerical methods. However, this work does not use any textual features specific to the task domain and focuses on different tokenization strategies. Unlike our LLMP-UCB algorithm, their prediction relies on obtaining the model's final output likelihoods. Nonetheless, their work describes patterns in time series that match the known capabilities of modern LLMs and how the composition of such patterns can present new challenges. Their observations on the repetition bias and arithmetic trends extrapolation informed the reward functions in~\ref{app:reward_function_ontology}.

\subsection{Combining Bandits and LLMs} 
The prevalence of foundation models trained on human-generated internet data has opened new avenues for tapping into inductive biases based on human notions of \textit{curiosity} and  \textit{exploration}~\citep{zhang2024omniopenendednessmodelshuman, faldor2025omniepicopenendednessmodelshuman}. The growing cost of training state-of-the-art (SOTA) API-served models like GPT-4o~\citep{openai2024gpt4ocard} is increasingly prohibitive due to chip infrastructure and staffing demands~\citep{cottier2025risingcoststrainingfrontier,li2023large}. This naturally calls for methods that leverage the capacity of already trained models through prompting and scaffolding algorithms.~\citet{bouneffouf2025surveymultiarmedbanditsmeet} point to this by emphasizing the importance of hybrid approaches in which numerical algorithms help refine prompts for LLMs, which then augment the context used by the bandits.~\citet{alamdari2024jumpstartingbanditsllmgenerated} focus on using bandits for recommendation systems. Their setting is also limited to language-based tasks where the LLM is leveraged to generate synthetic data to warm-start the context before deployment to real target users. More recently, \citet{lange2025shinkaevolve} use a UCB-style bandit to adaptively select which large language model in an ensemble proposes the most effective program mutation during an evolutionary optimization loop.

Close to our own framing is the concurrent work of \citet{gooding2025interaction}. Although they study LLM agents in a different setting, their vocabulary and framing were useful for articulating our experimental results in terms of embedding geometry.

\section{Reward Functions}
\label{app:reward_function_ontology}

This section outlines the details of different reward functions used to test the contextual bandits in the synthetic movie dataset experiment. 

\subsection{Piecewise Linear Numerical Reward Function}
\label{app:piece-wise-linear}

We define a parity-based reward function where the context consists of discrete user IDs and the reward depends on their modular arithmetic properties.

\textbf{Context Space:}
\begin{align}
x_t = \begin{bmatrix} 
\text{user\_id}_t \\ 
\text{genre}_t 
\end{bmatrix} \in \mathbb{Z}^2
\end{align}

where:
\begin{itemize}
\item $\text{user\_id}_t \in \mathbb{Z}^+$ represents the user identifier
\item $\text{genre}_t \in \mathbb{Z}^+$ represents the genre identifier
\end{itemize}

\textbf{Action Space:}
\begin{align}
\mathcal{A} = \mathbb{R}
\end{align}

\textbf{Parity Function:}
We define the parity function as:
\begin{align}
\text{parity}(n) = \begin{cases}
\text{odd} & \text{if } n \bmod 2 = 1 \\
\text{even} & \text{if } n \bmod 2 = 0
\end{cases}
\end{align}

\textbf{Reward Function:}
The parity-based reward function is defined as:
\begin{align}
r_t = f_{\text{parity}}(c_t, a_t) + \epsilon_t
\end{align}

where:
\begin{align}
f_{\text{parity}}(c_t, a_t) = \begin{cases}
4 \cdot a_t & \text{if } \text{parity}(\text{user\_id}_t) = \text{parity}(\text{genre}_t) \\
1 \cdot a_t & \text{if } \text{parity}(\text{user\_id}_t) \neq \text{parity}(\text{genre}_t)
\end{cases}
\end{align}

\textbf{Compact Representation:}
This can be expressed more compactly as:
\begin{align}
r_t = a_t \cdot (3 \cdot \mathbb{I}[\text{parity\_match}(c_t)] + 1) + \epsilon_t
\end{align}

where the parity matching indicator is:
\begin{align}
\text{parity\_match}(c_t) = \begin{cases}
1 & \text{if } (\text{user\_id}_t \bmod 2) = (\text{genre}_t \bmod 2) \\
0 & \text{otherwise}
\end{cases}
\end{align}

\newpage
\textbf{Properties of the reward function:}
\begin{itemize}
\item \textbf{Non-linear} in context features due to modular arithmetic operations
\item \textbf{Piecewise linear} in action with context-dependent multiplicative factors
\item \textbf{Discrete switching} behavior based on parity alignment between user and genre identifiers
\item \textbf{Higher reward multiplier} (4×) for parity-matched user-genre pairs
\item \textbf{Lower reward multiplier} (1×) for parity-mismatched pairs
\end{itemize}

% \newpage
\subsection{Non-Linear Numerical Reward Function}
\label{app:non_linear_numerical}

To test our architectures on a non-linear reward function, we define a parity-based reward function where the reward exhibits cubic and quadratic relationships with the action parameter.

\textbf{Context Space:}
\begin{align}
x_t = \begin{bmatrix} 
\text{user\_id}_t \\ 
\text{genre}_t 
\end{bmatrix} \in \mathbb{Z}^2
\end{align}

where:
\begin{itemize}
\item $\text{user\_id}_t \in \mathbb{Z}^+$ represents the user identifier
\item $\text{genre}_t \in \mathbb{Z}^+$ represents the genre identifier
\end{itemize}

\textbf{Action Space:}
\begin{align}
\mathcal{A} = \{0, 1, 2\}
\end{align}

\textbf{Parity Function:}
We define the parity function as:
\begin{align}
\text{parity}(n) = \begin{cases}
\text{odd} & \text{if } n \bmod 2 = 1 \\
\text{even} & \text{if } n \bmod 2 = 0
\end{cases}
\end{align}

\textbf{Reward Function:}
The polynomial parity-based reward function is defined as:
\begin{align}
r_t = f_{\text{poly-parity}}(c_t, a_t) + \epsilon_t
\end{align}

where:
\begin{align}
f_{\text{poly-parity}}(c_t, a_t) = \begin{cases}
a_t^3 \cdot \text{genre}_t & \text{if } \text{parity}(\text{user\_id}_t) = \text{parity}(\text{genre}_t) \\
a_t^2 \cdot \text{genre}_t & \text{if } \text{parity}(\text{user\_id}_t) \neq \text{parity}(\text{genre}_t)
\end{cases}
\end{align}

\textbf{Compact Representation:}
This can be expressed more compactly as:
\begin{align}
r_t = \text{genre}_t \cdot a_t^{2 + \mathbb{I}[\text{parity\_match}(c_t)]} + \epsilon_t
\end{align}

where the parity matching indicator is:
\begin{align}
\text{parity\_match}(c_t) = \begin{cases}
1 & \text{if } (\text{user\_id}_t \bmod 2) = (\text{genre}_t \bmod 2) \\
0 & \text{otherwise}
\end{cases}
\end{align}

Using indicator functions, we can write:
\begin{align}
r_t = \text{genre}_t \cdot \left[ \mathbb{I}[\text{parity\_match}(c_t)] \cdot a_t^3 + (1 - \mathbb{I}[\text{parity\_match}(c_t)]) \cdot a_t^2 \right] + \epsilon_t
\end{align}

\clearpage
\textbf{Properties of the reward function:}
\begin{itemize}
\item \textbf{Highly non-linear} in both context features and action parameters
\item \textbf{Polynomial dependence} on action with degree determined by parity matching
\item \textbf{Genre scaling} where the genre identifier acts as a multiplicative factor
\item \textbf{Cubic growth} ($a_t^3$) for parity-matched user-genre pairs
\item \textbf{Quadratic growth} ($a_t^2$) for parity-mismatched pairs
\item \textbf{Context-dependent polynomial degree} creating complex reward landscapes
\item \textbf{Asymmetric reward structure} with significantly different growth rates based on parity alignment
\end{itemize}

\textbf{Complexity Analysis:}
The reward function complexity increases significantly compared to linear models:
\begin{itemize}
\item For large positive actions: $|a_t| >> 1$, parity-matched pairs receive exponentially higher rewards
\item For small actions: $|a_t| < 1$, parity-mismatched pairs may receive higher rewards due to $a_t^2 > a_t^3$
\item The genre multiplier creates heterogeneous reward scales across different genre categories
\end{itemize}

\subsection{Highly Nonlinear Prime-Parity Reward Function}
\label{app:highly_non-linear}
Here we describe a reward function that incorporates nonlinear transformations of context and action features, as well as complex interactions based on number-theoretic properties (primality and parity).

\textbf{Context Space:}
\begin{align}
c_t = \begin{bmatrix} 
\text{user\_id}_t \\ 
\text{genre}_t 
\end{bmatrix} \in \mathbb{Z}^2
\end{align}

where:
\begin{itemize}
\item $\text{user\_id}_t \in \mathbb{Z}^+$ is the user identifier
\item $\text{genre}_t \in \mathbb{Z}^+$ is the genre identifier
\end{itemize}

\textbf{Action Space:}
\begin{align}
\mathcal{A} = \{0, 1, 2\}
\end{align}

\textbf{Auxiliary Functions:}
\begin{itemize}
    \item \textbf{Primality:} $\text{is\_prime}(n)$ returns $1$ if $n$ is prime, $0$ otherwise.
    \item \textbf{Parity:} $\text{is\_odd}(n)$ returns $1$ if $n$ is odd, $0$ otherwise.
\end{itemize}

\textbf{Reward Function:}
The reward function is defined as:
\begin{align}
r_t = f_{\text{nonlinear}}(c_t, a_t) + \epsilon_t
\end{align}

where:
\begin{align}
f_{\text{nonlinear}}(c_t, a_t) = \text{user\_effect} \cdot \text{genre\_effect} + \text{action\_effect} + \text{interaction} - 5 \cdot \mathbb{I}[a_t = 0]
\end{align}

with the following components:
\begin{align}
\text{user\_effect} &= \sin(\text{user\_id}_t) + \log(\text{user\_id}_t + 1) \\
\text{genre\_effect} &= \exp\left(\frac{\text{genre}_t}{5}\right) - \cos(\text{genre}_t) \\
\text{action\_effect} &= (a_t + 1)^2 \cdot \tan(a_t + 1)
\end{align}

\textbf{Interaction Term:}
\begin{align}
\text{interaction} = 
\begin{cases}
\sqrt{\text{user\_id}_t \cdot \text{genre}_t} \cdot (a_t + 1) & \text{if } \text{is\_prime}(\text{user\_id}_t) = 1 \text{ and } \text{is\_odd}(\text{genre}_t) = 1 \\
-\sqrt{\text{user\_id}_t + \text{genre}_t} \cdot (a_t + 1) & \text{if } \text{is\_prime}(\text{user\_id}_t) = 0 \text{ and } \text{is\_odd}(\text{genre}_t) = 0 \\
\log(\text{user\_id}_t + \text{genre}_t + a_t + 1) & \text{otherwise}
\end{cases}
\end{align}

\textbf{Penalty Term} is $-5 \cdot \mathbb{I} [ a_t = 0 ]$ where $\mathbb{I}[\cdot]$ is the indicator function.

\textbf{Properties:}
This reward function exhibits the following characteristics:
\begin{itemize}
    \item \textbf{Highly nonlinear} dependence on context and action features, including trigonometric, exponential, logarithmic, and polynomial terms.
    \item \textbf{Number-theoretic interactions:} The reward structure depends on whether the user ID is prime and whether the genre is odd.
    \item \textbf{Complex interaction term:} The interaction term can be positive, negative, or logarithmic, depending on the prime/parity status of the context.
    \item \textbf{Context-action coupling:} The reward is a nonlinear function of both context and action, with context-dependent polynomial and transcendental terms.
\end{itemize}

\textbf{Complexity Analysis:}
\begin{itemize}
    \item The reward function is not analytically tractable for optimization using standard linear or convex methods.
    \item The presence of trigonometric and exponential terms can lead to rapid changes in reward for small changes in context or action.
    \item The prime/parity interaction introduces discrete jumps in the reward surface, making exploration challenging.
\end{itemize}

\newpage
\subsection{LLM-as-a-Judge Reward Function}
\label{app:llm-as-a-judge}

Here we describe a reward function that leverages a Large Language Model (LLM) as an evaluator. The LLM interprets the textual context and the chosen action to directly generate a scalar reward signal based on a predefined rubric.

\textbf{Context Space}  We only use the natural language synopsis of the movie $\text{description}_t \in \mathcal{Z}$.

\textbf{LLM Specification:}
Let $f_{\text{LLM}}: \mathcal{Z} \times \mathcal{A} \rightarrow \mathcal{S}$ be the LLM generation function that maps a context-action pair to a structured string response $s_t$. The response includes a reasoning chain and a final score.

The specific instruction provided to the LLM enforces a rubric favoring ``comedic'' content based on the dataset in~\ref{app:dataset_gen}. The prompt structure is defined as follows:

\begin{tcolorbox}[colback=gray!10!white, colframe=black, title=Reward Prompt, boxrule=0.8pt, arc=4pt, fonttitle=\bfseries]
\begin{minted}[
    fontsize=\small,
    breaklines,
    breakanywhere,
    breaksymbolleft=,
    breaksymbolright=,
    breakautoindent=false
]{json}
{
    "role": "system",
    "content": "You are a helpful assistant that decides how to reward an action taken based on a movie description. The action can be either 'Serve' (indicating playing the movie preview), 'Recommend' (indicating showing the title but not playing it), or 'Do Not Recommend' (indicating that the movie should be ignored and not shown at all). Reward actions that recommend movies with comedic content. If a movie has a comedic description and the action is 'Recommend', give a reward of 1; if the action is 'Serve', give a reward of 2; if the action is 'Do Not Recommend', give a reward of 0. If a movie does not have a comedic description, give a reward of 2 for the 'Do Not Recommend' action and 0 for all other actions."
},
{
    "role": "user",
    "content": "### Description:\nA couple of people join a firefighting academy and go on to confront the culprits behind various arsons in their city, one of whom may be their former trainer.\n### Action:\nServe\n### Reasoning: This movie describes people fighting and solving mysteries in highly physically challenging settings, tropes that are commonly associated with action movies and not comedy. The action 'Serve' implies engaging with something that is not comedy. The reward is 0.\n\n...### Description: {description}\n### Action: {action}\n### Reasoning:"
}

\end{minted}
\end{tcolorbox}

\textbf{Action Space:}
\begin{align}
\mathcal{A} = \{ \text{Do Not Recommend}, \text{Recommend}, \text{Serve} \}
\end{align}

\textbf{Reward Function:}
The reward function is defined by parsing the generated output $s_t$:
\begin{align}
r_t = \text{ParseScore}(f_{\text{LLM}}(c_t, a_t))
\end{align}

where the rubric logic approximated by the LLM is:
\begin{align}
r_t \approx \begin{cases} 
2 & \text{if } \text{is\_comedy}(c_t) \land a_t = \text{Serve} \\
1 & \text{if } \text{is\_comedy}(c_t) \land a_t = \text{Recommend} \\
2 & \text{if } \neg\text{is\_comedy}(c_t) \land a_t = \text{Do Not Recommend} \\
0 & \text{otherwise}
\end{cases}
\end{align}

\newpage
\textbf{Properties:}
\begin{itemize}
    \item \textbf{Black-box mapping:} The reward is a non-linear, complex function of the input text, derived from the LLM's internal knowledge base.
    \item \textbf{Instruction-following:} The reward logic is programmable via natural language prompts rather than mathematical weights.
    \item \textbf{Zero-shot capability:} Can evaluate complex criteria (e.g., ``comedic'', ``family-friendly'') without training a specific feature extractor.
    \item \textbf{Reasoning trace:} The LLM provides textual justification (Chain-of-Thought) alongside the numerical score.
\end{itemize}

\textbf{Complexity Analysis:}
\begin{itemize}
    \item High computational cost due to inference time for each reward calculation ($O(\text{model\_depth} \times \text{seq\_len}^2)$).
    \item High latency, making it less suitable for real-time loops compared to feature-extracted rewards.
\end{itemize}

\subsection{Feature-Extracted Reward Function}
\label{app:feature-extraction-based-reward}
Here we describe a reward function that leverages features extracted from textual context, such as readability and sentiment, to determine the reward for a given action.

\textbf{Context Space:}
\begin{align}
c_t = \begin{bmatrix} 
\text{user\_id}_t \\ 
\psi(\text{description}_t) 
\end{bmatrix}
\end{align}

where:
\begin{itemize}
\item $\text{user\_id}_t \in \mathbb{Z}^+$ is the user identifier
\item $\text{description}_t \in \mathcal{Z}$ is a natural language description
\end{itemize}

\textbf{Feature Extraction:}
Let $\psi: \mathcal{Z} \rightarrow \mathbb{R}^3$ be a feature extraction function that maps the description to a vector of features:
\begin{align}
\psi(\text{description}_t) = \begin{bmatrix}
\text{readability}_t \\
\text{polarity}_t \\
\text{subjectivity}_t
\end{bmatrix}
\end{align}

where:
\begin{itemize}
    \item $\text{readability}_t$ is the Flesch Reading Ease score (scaled to $[0,100]$)
    \item $\text{polarity}_t$ is the sentiment polarity (range $[-1,1]$)
    \item $\text{subjectivity}_t$ is the sentiment subjectivity (range $[0,1]$)
\end{itemize}

\textbf{Action Space:}
\begin{align}
\mathcal{A} = \{0, 1, 2\}
\end{align}
where
\begin{itemize}
    \item $0$: Do not Recommend
    \item $1$: Recommend
    \item $2$: Serve
\end{itemize}

\newpage
\textbf{Reward Function:}
The reward function is defined as:
\begin{align}
r_t = f_{\text{features}}(c_t, a_t) + \epsilon_t
\end{align}

where the feature score is a weighted sum:
\begin{align}
\text{feature\_score}_t = 0.2 \cdot \left(\frac{\text{readability}_t}{100}\right) + 0.4 \cdot \text{polarity}_t + 0.4 \cdot \text{subjectivity}_t
\end{align}

and the reward mapping is:
\begin{align}
f_{\text{features}}(c_t, a_t) = 
\begin{cases}
\text{feature\_score}_t & \text{if } a_t = 1 \\
1.5 \cdot \text{feature\_score}_t & \text{if } a_t = 2 \\
-\text{feature\_score}_t & \text{if } a_t = 0 \\
0 & \text{otherwise}
\end{cases}
\end{align}

\textbf{Properties:}
\begin{itemize}
    \item \textbf{Linear mapping:} The reward is a linear function of extracted features, with action-dependent scaling.
    \item \textbf{Textual context:} The reward depends on natural language features, making it sensitive to the content and style of the description.
    \item \textbf{Action-dependent scaling:} The action determines both the sign and magnitude of the reward.
    \item \textbf{Interpretability:} The weights on readability, polarity, and subjectivity allow for transparent analysis of feature contributions.
    \item \textbf{Hybrid context:} Combines numerical (user ID) and textual (description) context, with the latter mapped to feature space.
\end{itemize}

\textbf{Complexity Analysis:}
\begin{itemize}
    \item The reward function is piecewise linear in the extracted features.
    \item The use of natural language feature extraction introduces nontrivial dependencies on the description, but the final mapping is simple and interpretable.
\end{itemize}

\newpage
\clearpage
\section{Numerical Bandit Implementation}
\label{app:numerical_baselines}

The first step is to establish reliable baselines and a unified training interface that passes both the programmatic and algorithmic sanity checks. 

We start by implementing an environment with the OpenAI Gym interface and test both a simple linear lookup reward function and one where the context-action pairs are non-linearly related to the reward. 

\begin{figure}[h!]
    \centering
    \begin{subfigure}[t]{0.47\textwidth}
        \centering
        \includegraphics[width=\linewidth]{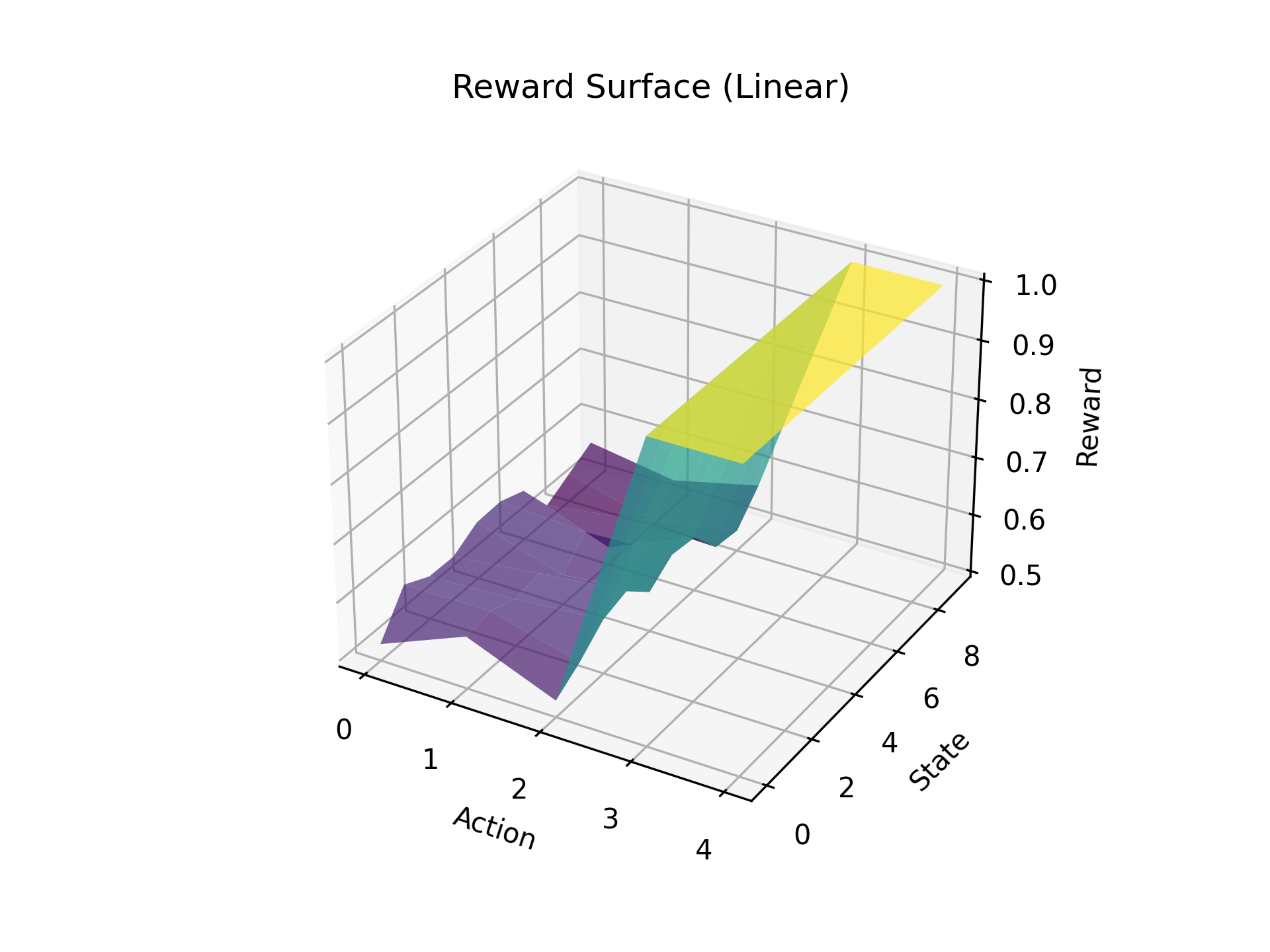}
        \caption{Linear Reward function}
        \label{fig:reward_mesh_linear}
    \end{subfigure}
    \hfill
    \begin{subfigure}[t]{0.47\textwidth}
        \centering
        \includegraphics[width=\linewidth]{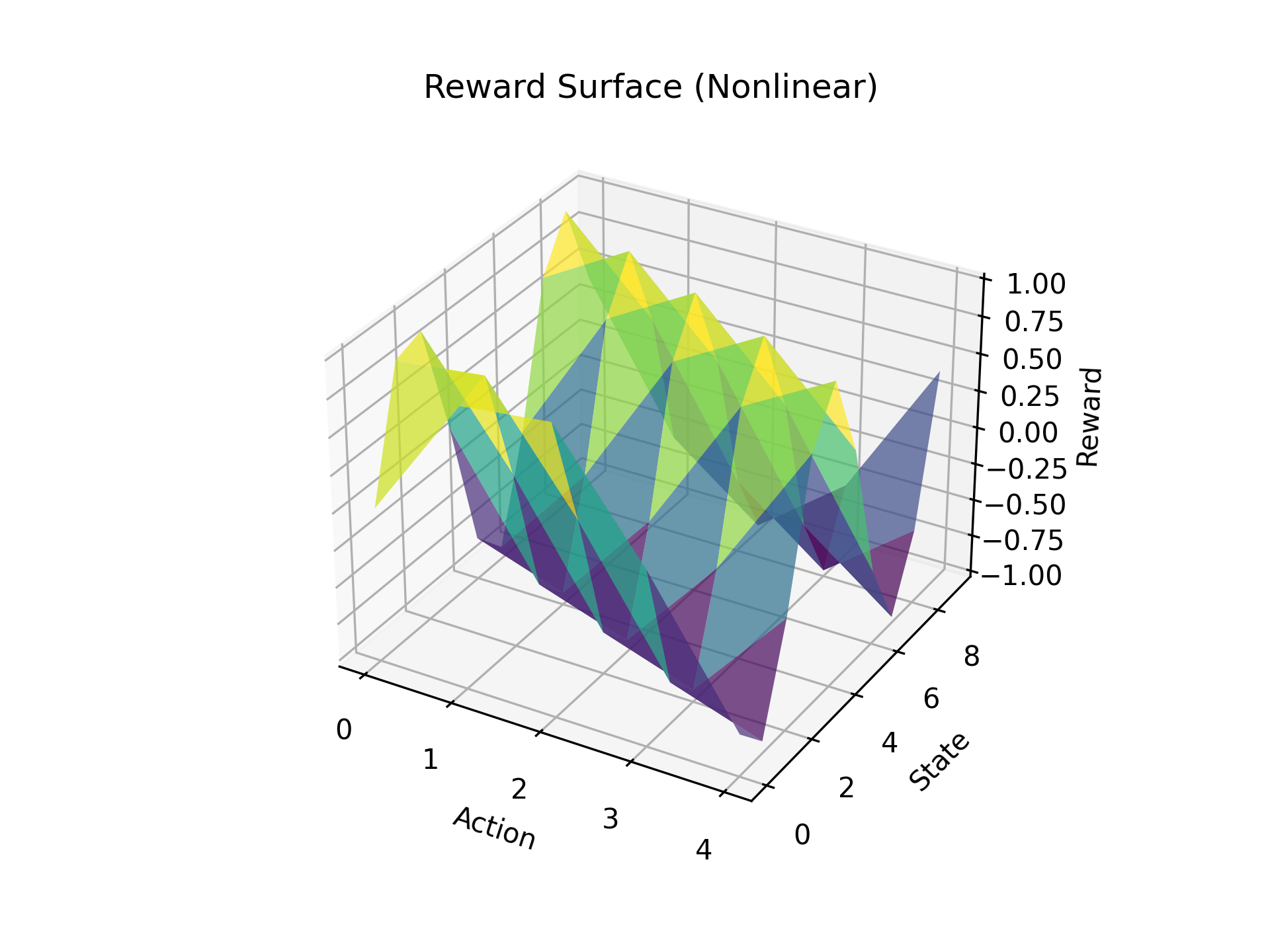} 
        \caption{Sinusoidal reward function}
        \label{fig:reward_mesh_nonlinear}
    \end{subfigure}
    \caption{Mesh plot of the reward functions}
    \label{fig:reward_meshes}
\end{figure}

\newpage
\subsection{Numerical Bandit Results}

Here we measure the performance on contextual multi-armed bandits algorithms. Specifically we test Thompson Sampling, Gaussian Process Bandits and LinearUCB. The simple test environment has a reward lookup, pre-determined offsets and optimal arms. The environment is fully stationary. The red dots show the bandit's pulls with the radius being proportional to the frequency that arm was chosen in that context and received the reward determined by the reward function.

\begin{figure}[h!]
    \centering
    \begin{subfigure}[t]{0.45\textwidth}
        \centering
        \includegraphics[width=\linewidth]{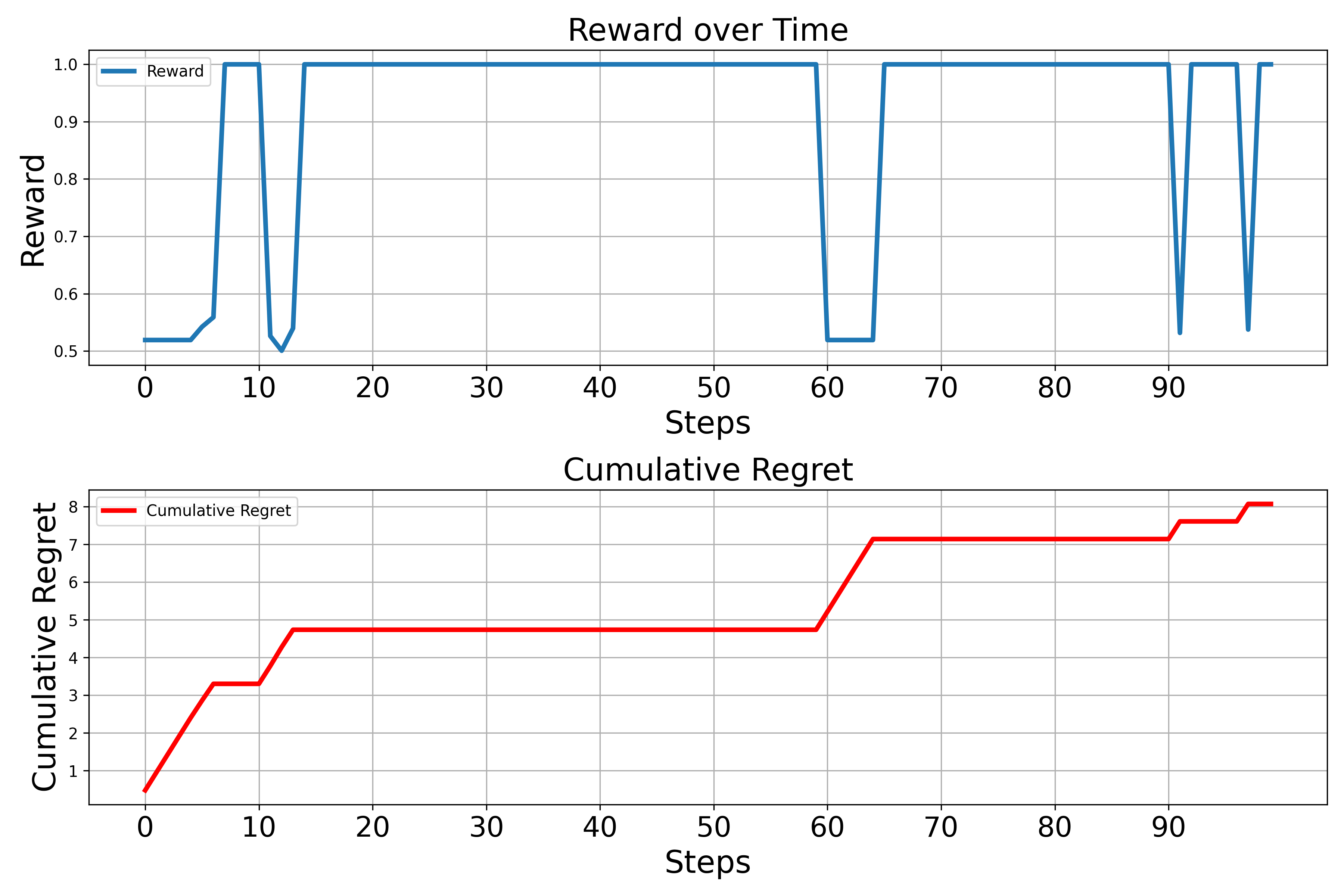}
    \end{subfigure}
    \hfill
    \begin{subfigure}[t]{0.45\textwidth}
        \centering
        \includegraphics[width=\linewidth]{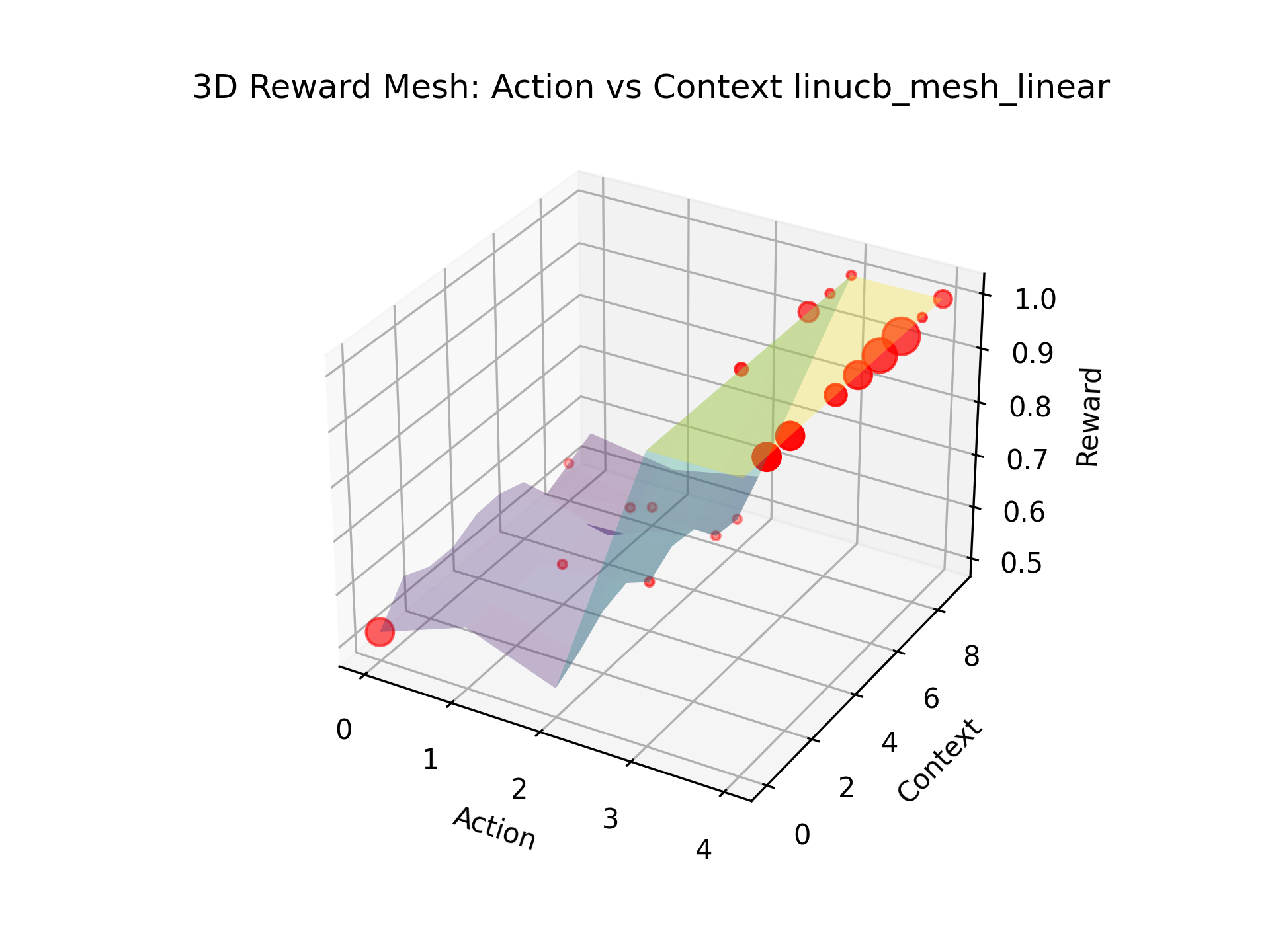}
    \end{subfigure}
    \caption{LinUCB performance with a linear reward function}
    \label{fig:linUCB-linear}
\end{figure}

\begin{figure}[h]
    \centering
    \begin{subfigure}[t]{0.45\textwidth}
        \centering
        \includegraphics[width=\linewidth]{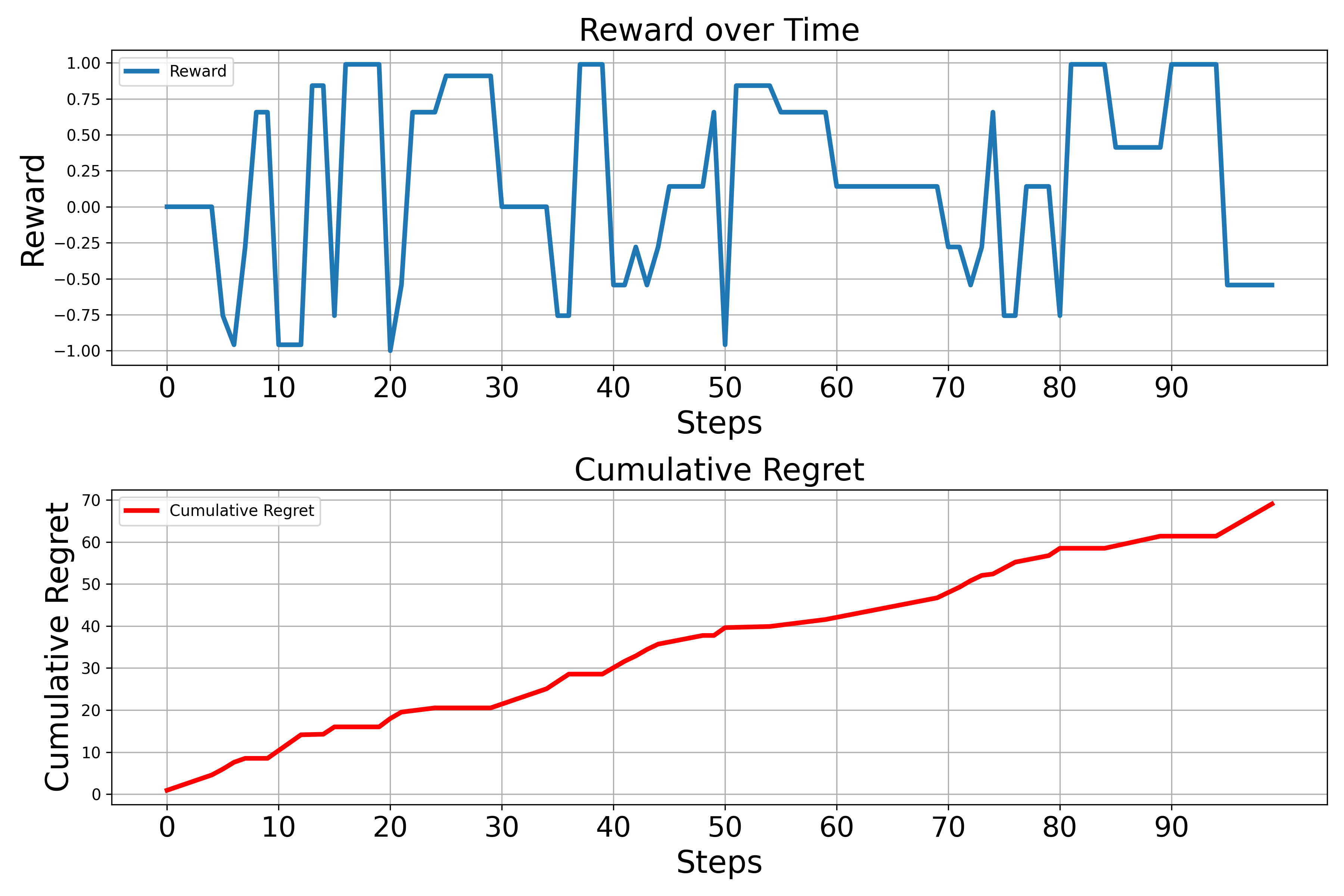}
    \end{subfigure}
    \hfill
    \begin{subfigure}[t]{0.45\textwidth}
        \centering
        \includegraphics[width=\linewidth]{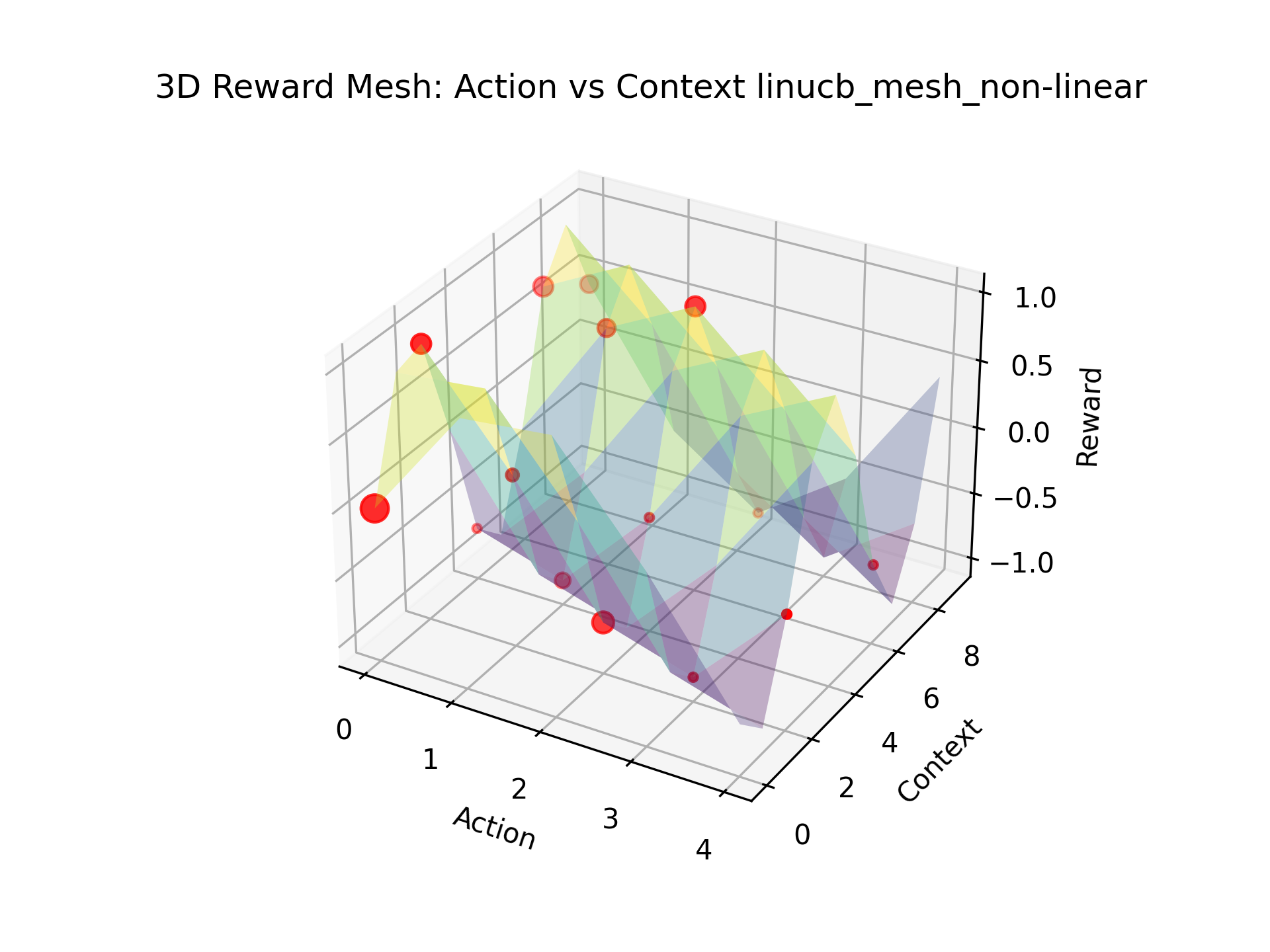}
    \end{subfigure}
    \caption{LinUCB performance with a non-linear reward function}
    \label{fig:linUCB-linear}
\end{figure}

\begin{figure}[h]
    \centering
    \begin{subfigure}[t]{0.45\textwidth}
        \centering
        \includegraphics[width=\linewidth]{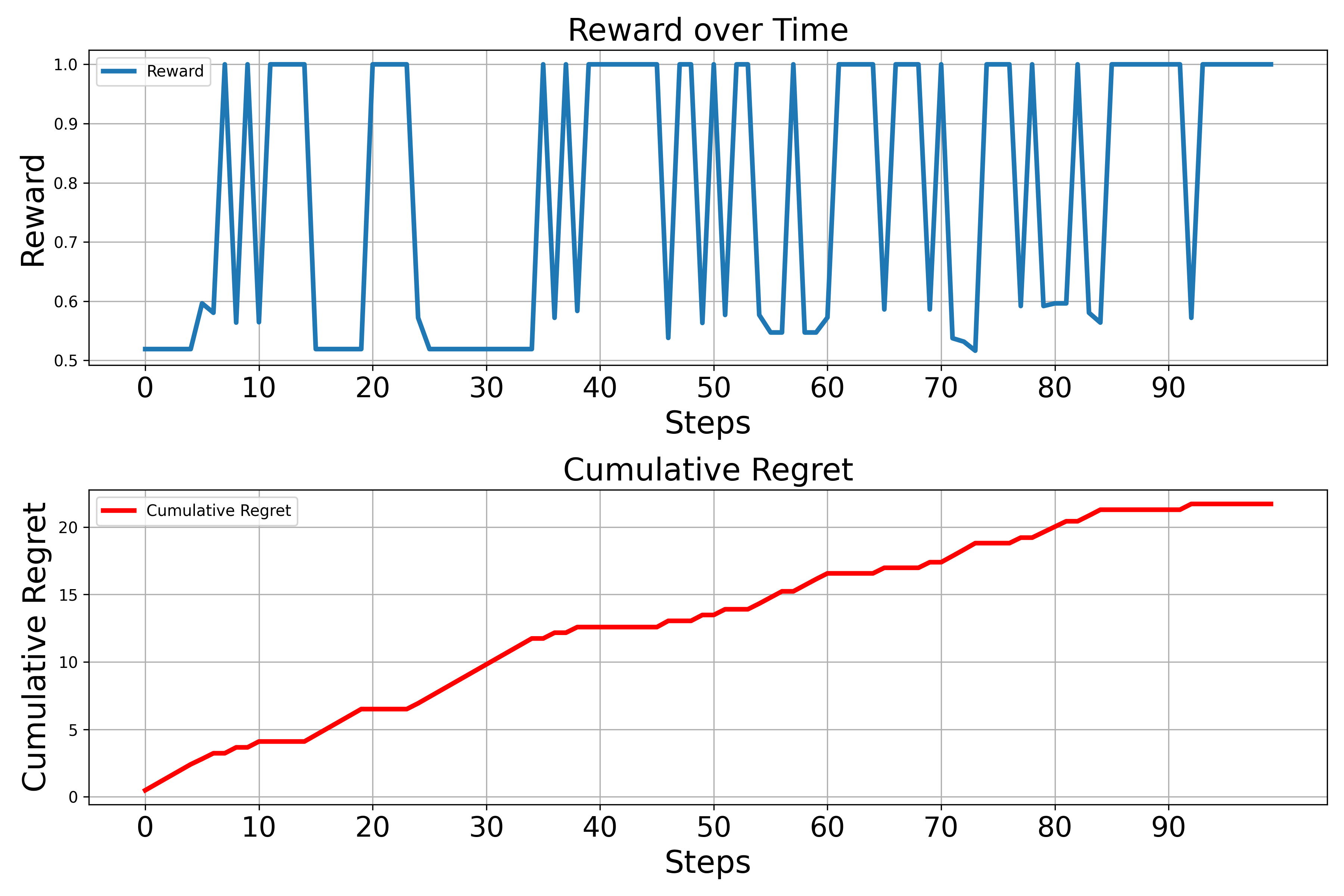}
    \end{subfigure}
    \hfill
    \begin{subfigure}[t]{0.45\textwidth}
        \centering
        \includegraphics[width=\linewidth]{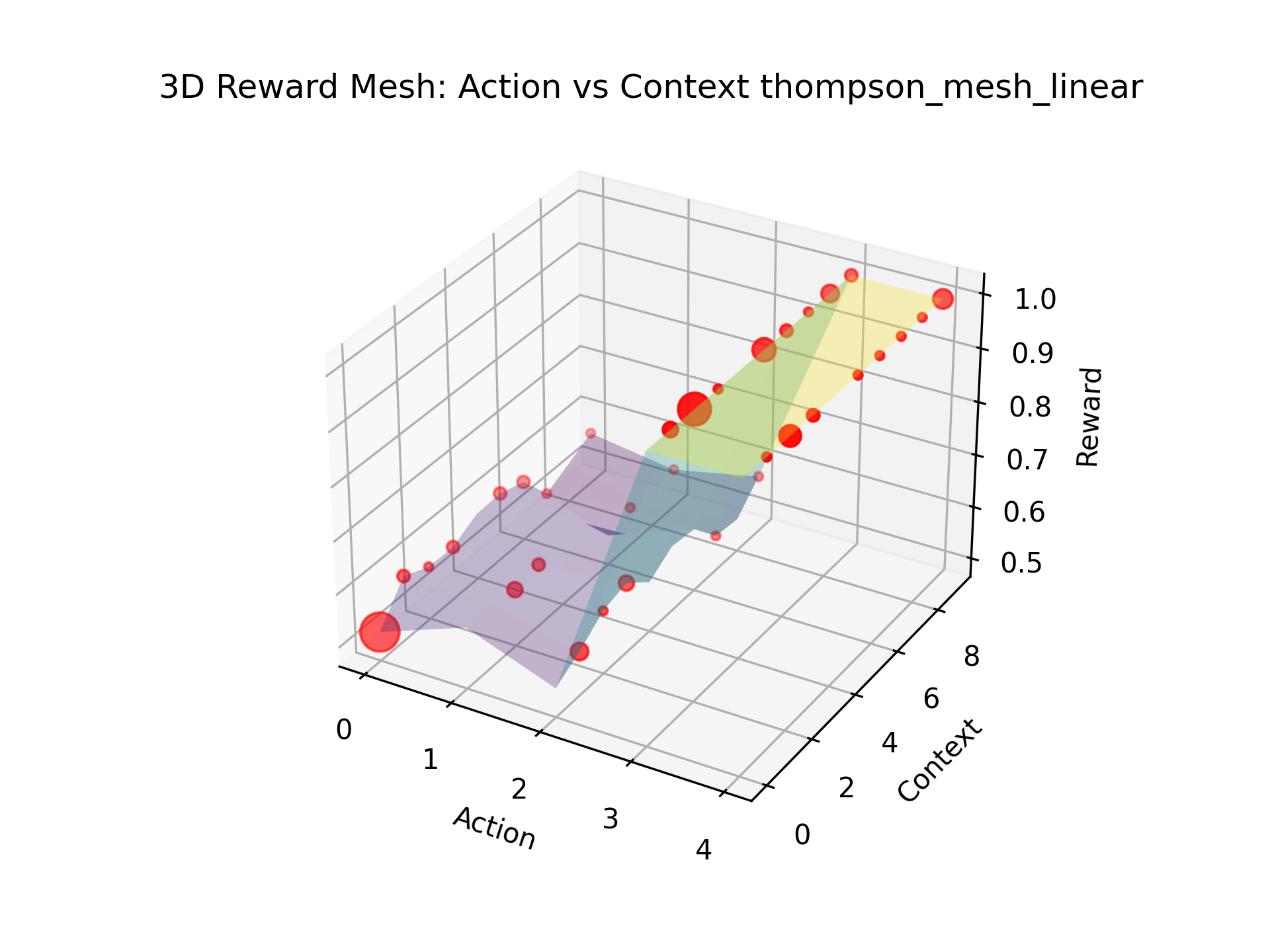}
    \end{subfigure}
    \caption{Thompson performance with a linear reward function}
    \label{fig:linUCB-linear}
\end{figure}

\begin{figure}[h]
    \centering
    \begin{subfigure}[t]{0.45\textwidth}
        \centering
        \includegraphics[width=\linewidth]{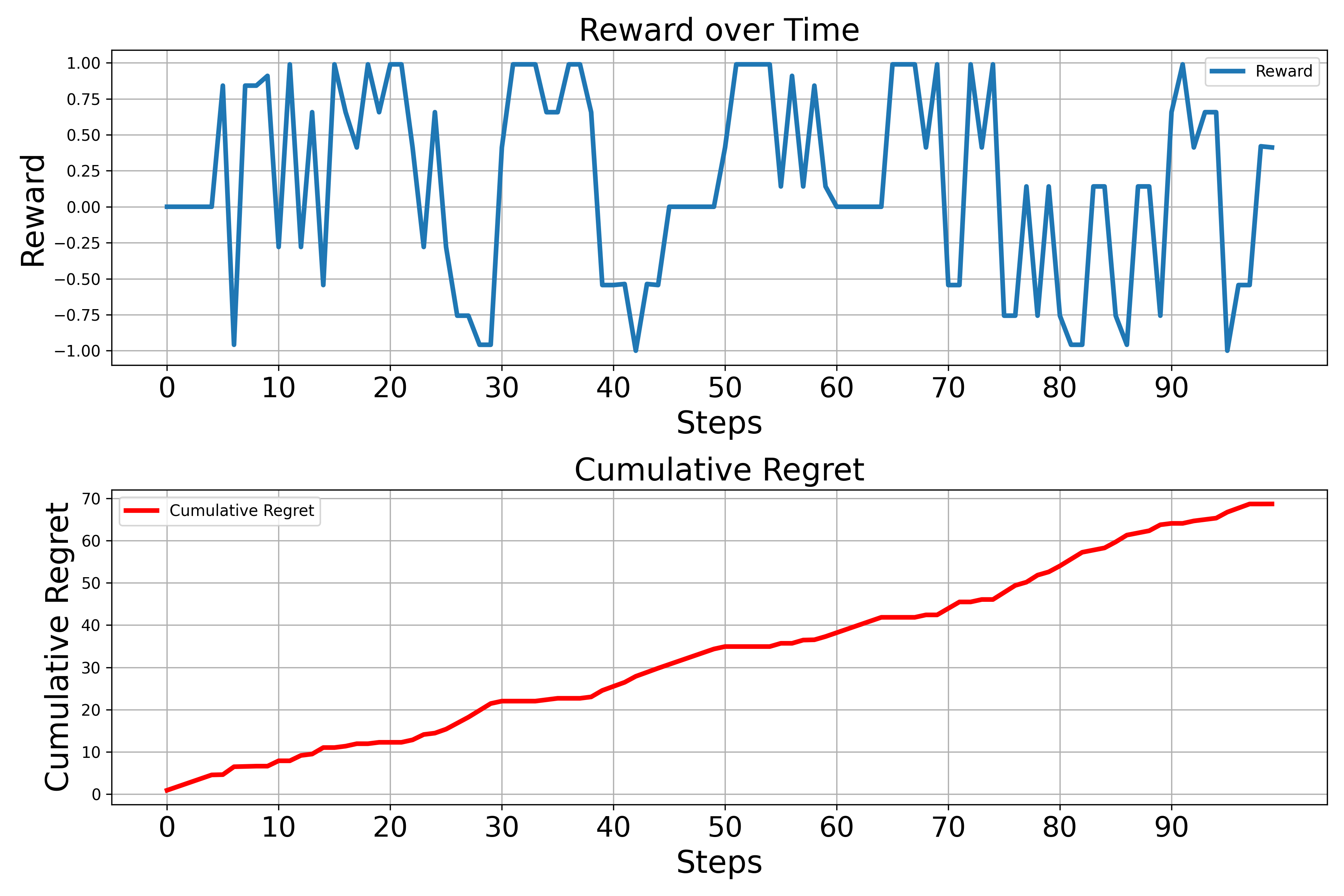}
    \end{subfigure}
    \hfill
    \begin{subfigure}[t]{0.45\textwidth}
        \centering
        \includegraphics[width=\linewidth]{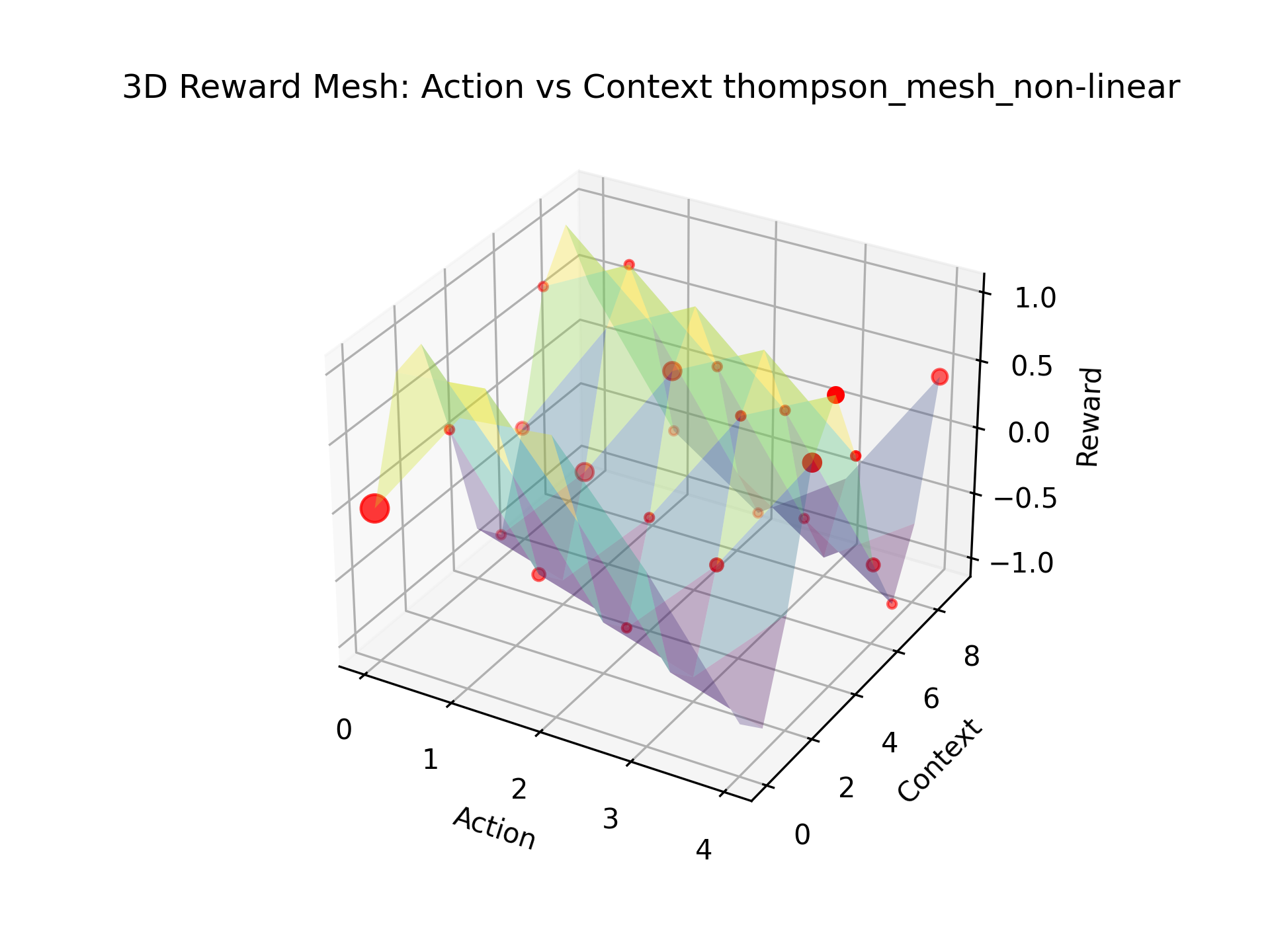}
    \end{subfigure}
    \caption{Thompson performance with a non-linear reward function}
    \label{fig:linUCB-linear}
\end{figure}

\begin{figure}[h]
    \centering
    \begin{subfigure}[t]{0.45\textwidth}
        \centering
        \includegraphics[width=\linewidth]{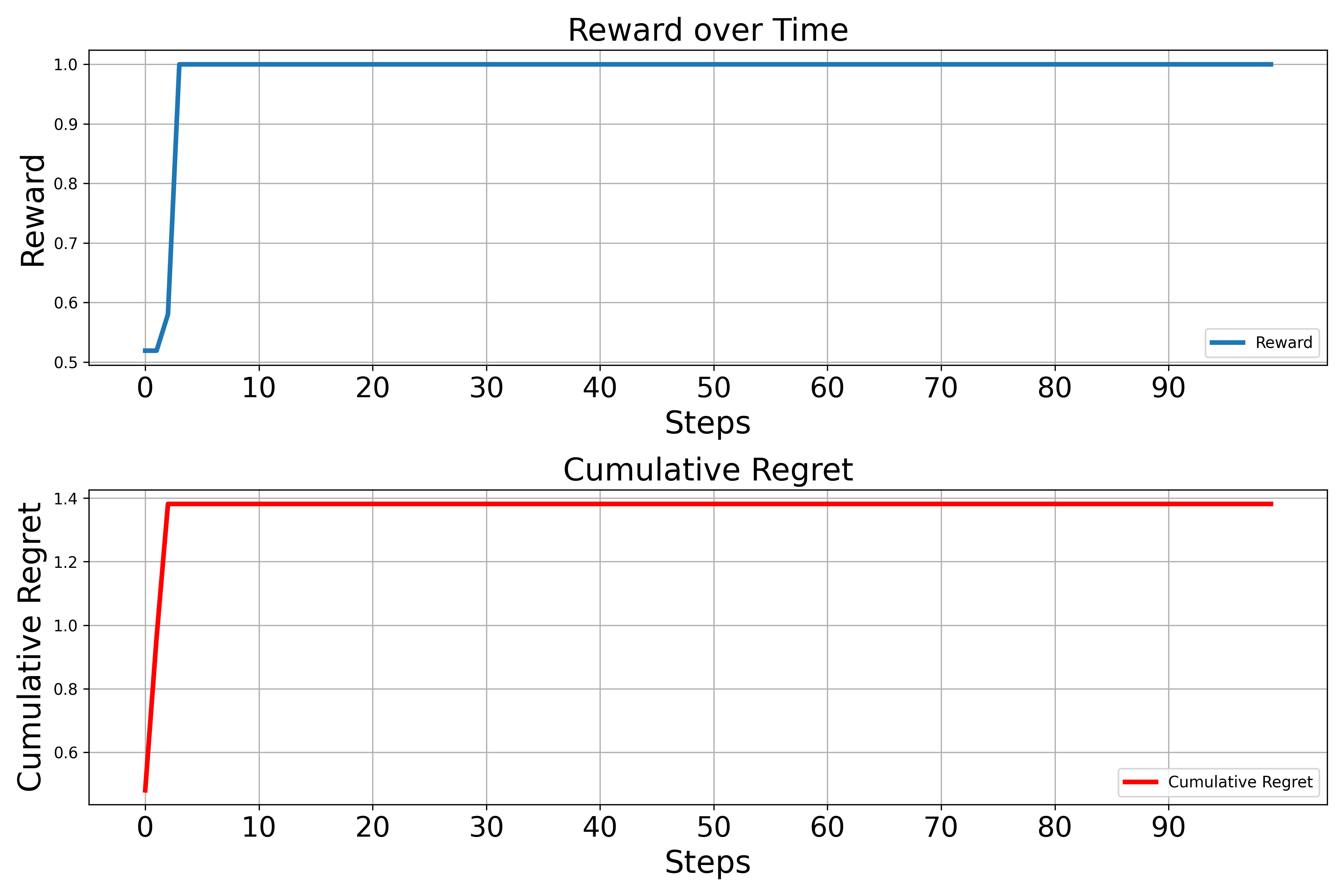}
    \end{subfigure}
    \hfill
    \begin{subfigure}[t]{0.45\textwidth}
        \centering
        \includegraphics[width=\linewidth]{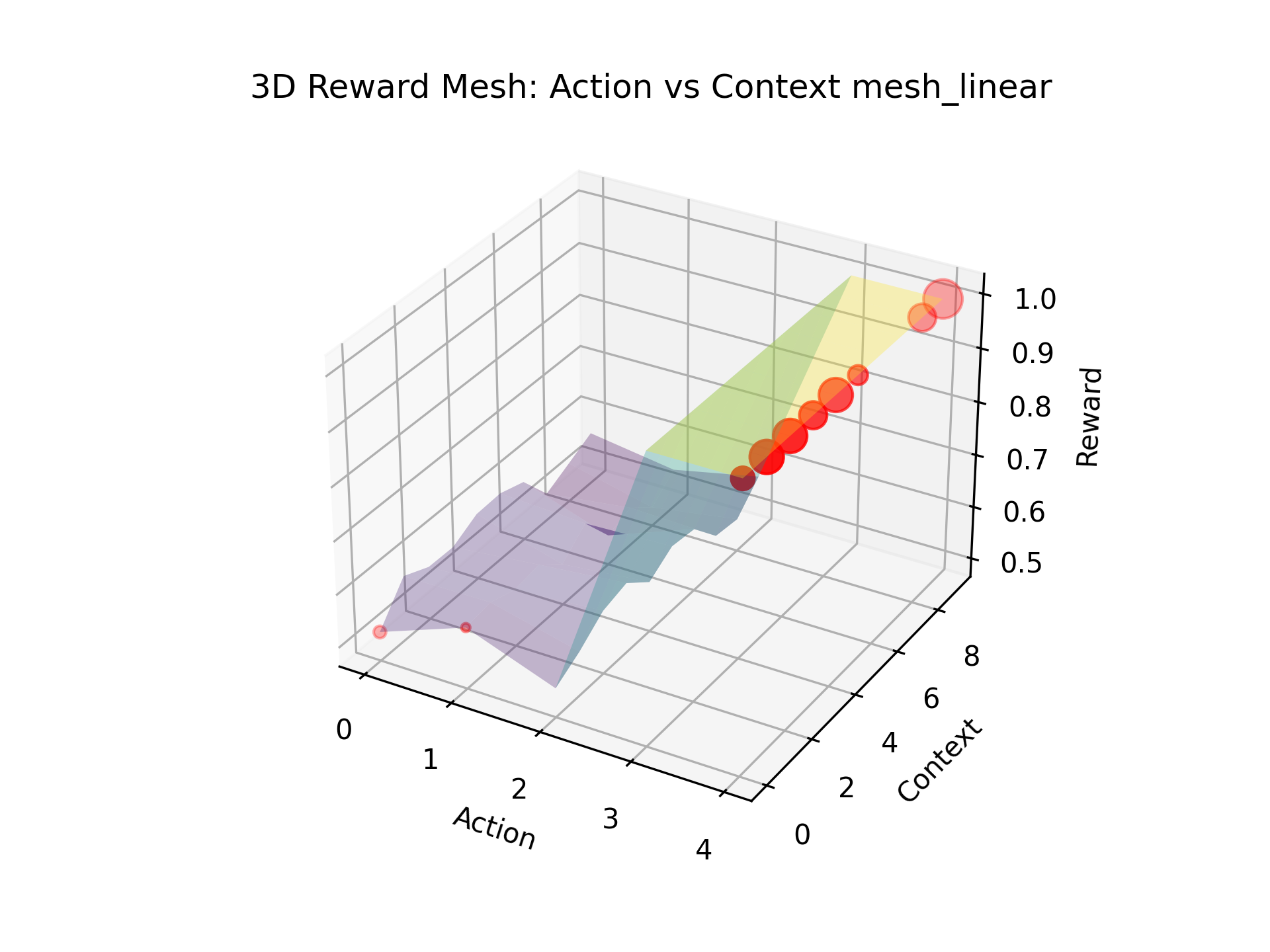}
    \end{subfigure}
    \caption{GPUCB performance with a linear reward function}
    \label{fig:linUCB-linear}
\end{figure}

\begin{figure}[h]
    \centering
    \begin{subfigure}[t]{0.45\textwidth}
        \centering
        \includegraphics[width=\linewidth]{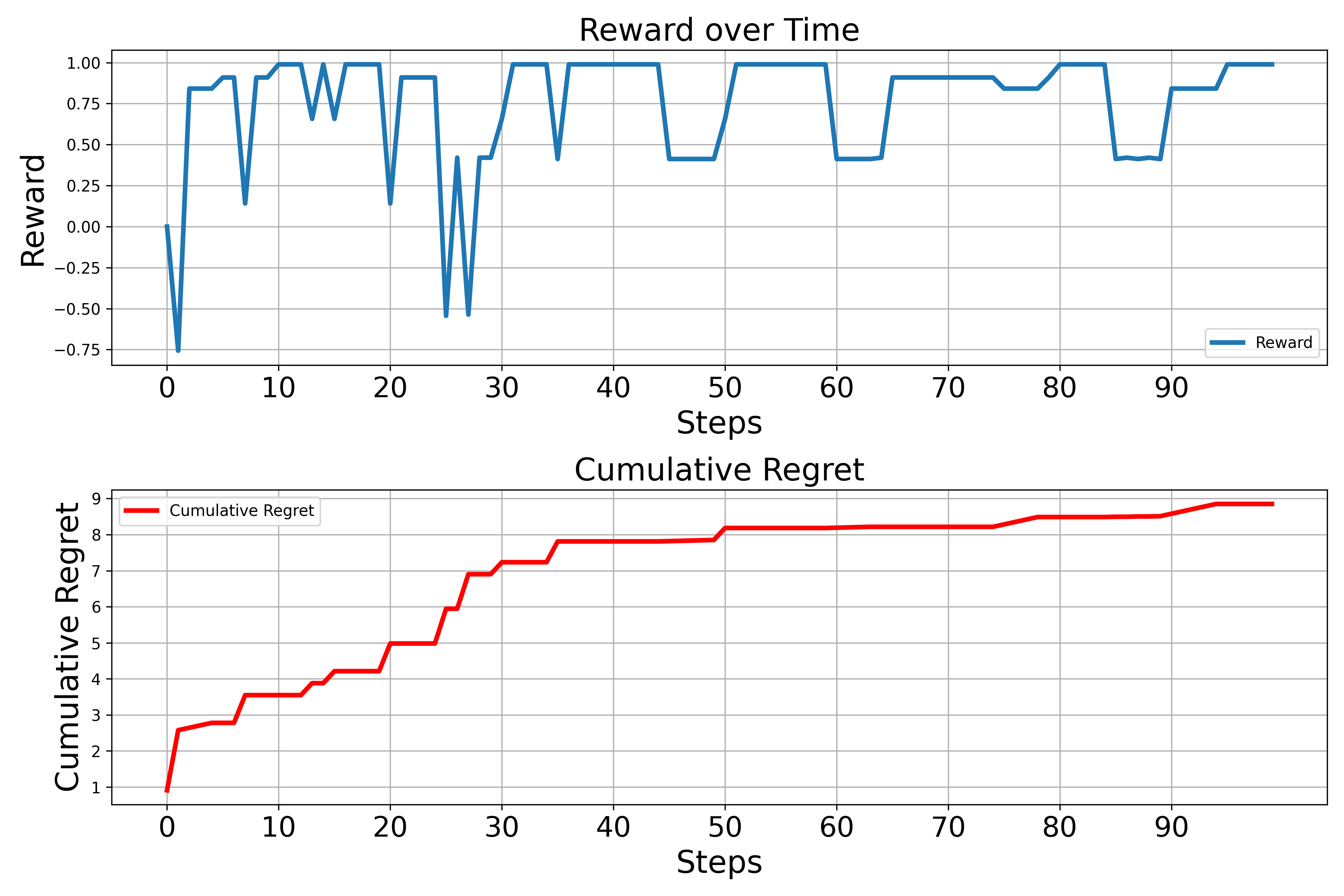}
    \end{subfigure}
    \hfill
    \begin{subfigure}[t]{0.45\textwidth}
        \centering
        \includegraphics[width=\linewidth]{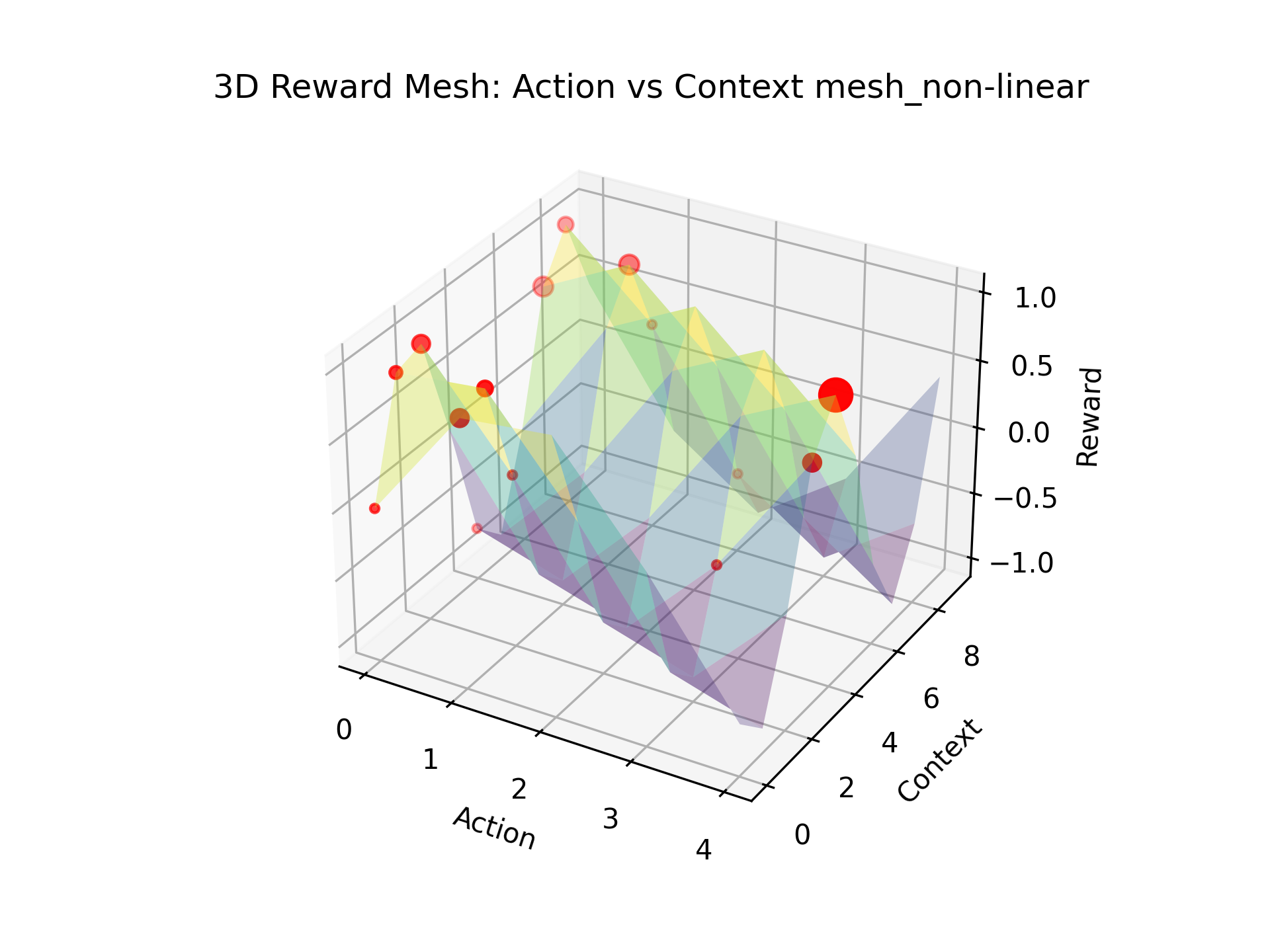}
    \end{subfigure}
    \caption{GPUCB performance with a non-linear reward function}
    \label{fig:linUCB-linear}
\end{figure}

\clearpage

\section{Dataset Generation}

\label{app:dataset_gen}
While the main goal of our work is to apply the method to public benchmarks, we first generate a set of context instances $\{c\}$ and possible actions $\{a\}$. We start by defining the action space $\mathcal{A}$ and the numerical and natural language features of the context $\mathcal{C}$. We choose a discrete set of actions which are relevant to the problem setting.

Then, the dataset is generated by asking \texttt{Claude Sonnet 4} to generate examples given the action and context space definition.

\textbf{Movie Watcher Dataset:}
\begin{itemize}
    \item \textbf{Numerical Features:} User ID, Genre category
    \item \textbf{Text Features:} Movie Plot Description.
    \item \textbf{Arms:} 
    \begin{itemize}
        \item Do Not Recommend (movie is not shown to user),
        \item Recommend (the movie is shown to the user in a list),
        \item Serve (the movie is shown and starts auto-playing).
    \end{itemize}
\end{itemize}

Once the full context dataset is generated, $ \forall c_i \in \mathcal{C}, \forall a_j \in \mathcal{A}$, we evaluate $f_{\mathrm{LLM}}(c_i, a_j)$ as described in~\autoref{app:llm-as-a-judge}.

\begin{tcolorbox}[colback=gray!10!white, colframe=black, title=Movie Dataset Sample, boxrule=0.8pt, arc=4pt, fonttitle=\bfseries]
\begin{minted}[
    fontsize=\small,
    breaklines,
    breakanywhere,
    breaksymbolleft=,
    breaksymbolright=,
    breakautoindent=false
]{json}
{   "user_id": 1,
    "genre": 1,
    "description": "A retired spy is forced back into action to save the city from a mysterious villain.",
    "action": 0,
    "ground_reward": 0},
{
    "user_id": 1,
    "genre": 1,
    "description": "A retired spy is forced back into action to save the city from a mysterious villain.",
    "action": 1,
    "ground_reward": 1},
{
    "user_id": 1,
    "genre": 1,
    "description": "A retired spy is forced back into action to save the city from a mysterious villain.",
    "action": 2,
    "ground_reward": 2},
{
    "user_id": 2,
    "genre": 2,
    "description": "Two best friends open a bakery and get into hilarious mishaps with their quirky customers.",
    "action": 0,
    "ground_reward": 0},
\end{minted}
\end{tcolorbox}

\newpage

\section{Assets and Licenses}
\label{app:embeddings_model_assets}

\subsection{Embeddings Models}
\begin{table*}[h]
\caption{Embeddings Models used}
\label{tab:assets}
\centering
\begin{tabular}{lcccccc}
\toprule           
\textbf{Asset} & \textbf{License} & \textbf{URL} \\
\midrule
\texttt{Dense Embeddings} &  Apache 2.0 & \small{\href{https://huggingface.co/NeuML/pubmedbert-base-embeddings}{\texttt{NeuML/pubmedbert-base-embeddings}}}  \\
\texttt{Matryoshka Embeddings} & Apache 2.0 & \small{\href{https://huggingface.co/NeuML/pubmedbert-base-embeddings-matryoshka}{\texttt{NeuML/pubmedbert-base-embeddings-matryoshka}}}  \\

\end{tabular}
\end{table*}

\subsection{Dataset Licenses}
\label{app:licenses_assets}

The licenses of the dataset used in this work that were not generated using a commercial language model.

\begin{table*}[h]
\caption{Datasets used}
\label{tab:dataset_licenses}
\centering
\begin{tabular}{lcccccc}
\toprule           
\textbf{Asset} & \textbf{License} & \textbf{URL} \\
\midrule
\texttt{Banking77} &  Apache 2.0 & \small{\href{https://huggingface.co/datasets/PolyAI/banking77}{\texttt{PolyAI/banking77}}}  \\
\texttt{TREC} & Apache 2.0 & \small{\href{https://huggingface.co/datasets/CogComp/trec}{\texttt{CogComp/trec}}}  \\

\end{tabular}
\end{table*}

\end{document}